\documentclass[11pt,a4paper]{article}

\usepackage{liatuo}          
\usepackage{multicol}
\usepackage{adjustbox}
\usepackage{makecell}
\usepackage{multirow}
\usepackage{placeins}
\usepackage{flafter}
\usepackage{dblfloatfix}
\title{\Large ME-VLM: A Unified VLM for Embodied Cognition and \\ Agent Coordination}
\author{\large \textbf{Foundation Model, Li Auto Inc.}}

\begin{document}

\makeLiTitle

\phantomsection
\begin{liabstract}
Physical AI requires models to ground visual and linguistic understanding in real-world environments while accounting for environmental constraints and execution feedback.
We introduce MachEmbodied-VLM (ME-VLM), a unified vision-language model with two variants, 4B and 35B-A3B, that brings together embodied cognition and multimodal agent capabilities.
Our work emphasizes physical perception and spatiotemporal reasoning, together with planning, interaction, and outcome assessment in both digital and physical environments.
We construct training data spanning embodied and multimodal agent tasks, including execution observations and feedback to support outcome assessment and decision refinement.
The training pipeline comprises embodied capability injection, separate reinforcement learning of embodied and multimodal-agent experts, and multi-teacher on-policy distillation that consolidates their complementary capabilities into a single model.
Experiments show competitive performance on both embodied and agent benchmarks, as well as on autonomous-driving and embodied-navigation tasks.
For edge deployment, visual token compression, W4A8 quantization, and hardware--software co-optimization enable on-device inference of the 4B variant on the M100, reducing prefill latency from 400~ms to 188~ms.\\[4pt]

\noindent\textit{\textbf{Project Page:}\enspace\url{https://machembodied.com/ME-Brain/ME-VLM.html}} 

\noindent\textit{\textbf{Code Repository:}\enspace\url{https://github.com/MachEmbodied/ME-VLM}}

\end{liabstract}

\begin{figure}[!htbp]
    \centering
    \includegraphics[width=\linewidth]{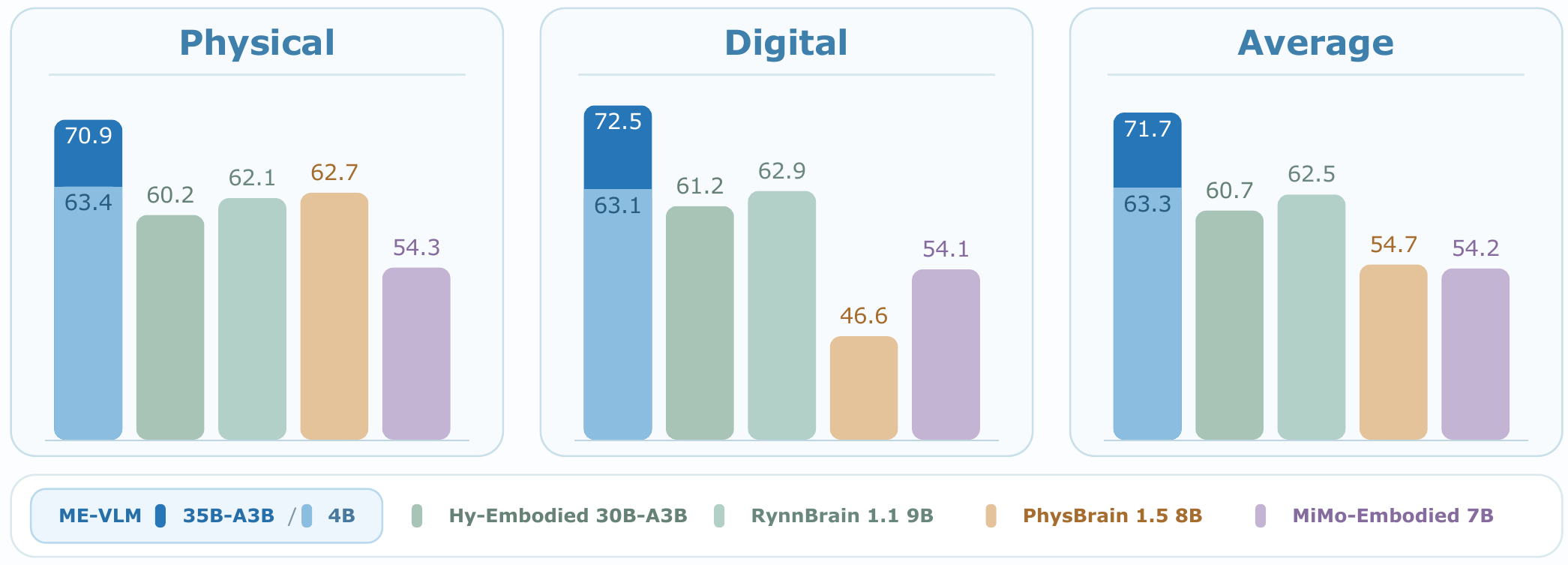}
    \caption{\textbf{Performance Comparison.} ME-VLM (35B-A3B) outperforms strong embodied VLM baselines of comparable scale on embodied and agent benchmarks.}
    \label{fig:abstract:overall_comparison}
\end{figure}

\clearpage

\tableofcontents

\clearpage


\section{Introduction}
\label{sec:introduction}

In recent years, digital agents \cite{qwen3_5, kimi_k3, glm5, team2026mach} have expanded from digital environments to the physical world. Driven by continued progress in large vision-language models, digital agents can now interpret complex instructions, plan long-horizon tasks, invoke external tools, and adjust their behavior according to feedback. Their success in digital interaction scenarios points to their potential to serve as a general-purpose ``brain'' in the real world. In Physical AI, however, language and visual reasoning alone cannot support reliable real-world interaction: an agent should translate high-level semantic goals into physical behaviors aligned with the real-world environments while perceiving environmental changes, evaluating the effects of its actions, and revising subsequent decisions throughout execution.

These observations lead us to argue that an embodied agent operating in the physical world requires at least two synergistic core capabilities. The first is embodied cognition: the ability to perceive, understand, and reason about physical environments, with the key breakthroughs being strengthened geometric perception of physical elements and space, modeling of temporal action sequences, and modeling of execution feedback, which together establish native cognition of the real physical world. 
The second is agent capability: building on the enhanced spatial, temporal, and dynamics awareness, the agent combines long-horizon autonomous planning, skill and tool invocation, and reflection with failure attribution to drive its embodied platform through operations in real physical scenes, achieving a closed loop from perception and planning to action execution, feedback, and correction. An agent equipped with both capabilities could turn an open-ended goal into an executable process spanning multiple steps and skills, adapt its policy when the environment changes or execution fails, and thereby be deployed in real-world industries, robotics, and intelligent manufacturing.

Guided by this capability framework, this paper proposes MachEmbodied-VLM (ME-VLM), a foundation model for physical AI that unifies embodied cognition and agent capabilities. To this end, we
construct multi-dimensional training data covering both capability domains and design a unified joint training pipeline, enabling the model to learn embodied cognition and agent capabilities
within a single framework. Unlike existing pipelines that simply cascade a planner and executors \cite{black2025pi05}, ME-VLM adopts a unified modeling approach in which embodied cognition and agentic decisions
mutually constrain each other: perception and understanding of the real physical environment ground task planning and skill selection, while the physical states observed during execution in turn steer the model's attention toward the relevant objects and target regions and inform its subsequent decisions.

As shown in Figure~\ref{fig:abstract:overall_comparison}, ME-VLM achieves competitive performance on both embodied benchmarks and agent benchmarks. Beyond these settings, ME-VLM can be adapted to autonomous driving and embodied navigation, attaining competitive results on driving and navigation benchmarks as well, indicating that the learned embodied capabilities have potential for transfer across tasks.

Considering the stringent requirements on inference latency, power consumption, and compute in real-world deployments, we further co-optimize ME-VLM for edge-side chips. On the algorithmic side, we compress non-essential inference tokens, reduce redundant intermediate reasoning, and lower the frequency of model invocations during task execution, thereby reducing the agent's inference overhead. On the system side, quantized deployment, operator adaptation, and hardware–software co-optimization improve on-device execution efficiency. Together, these optimizations enable ME-VLM to perform low-latency local inference under limited compute resources, supporting the real-time execution loop of embodied systems.

In summary, our main contributions are as follows:
  \begin{itemize}
      \item \textbf{An embodied VLM with enhanced physical cognition.} Building on strengthened physical cognition, ME-VLM realizes a complete closed loop spanning user intent understanding, environment perception, and task planning, through to skill invocation, action execution, and outcome verification and correction.
      \item \textbf{Unified training data and a joint training pipeline.}  We construct training data covering physical understanding, task planning, skill invocation, execution-process understanding, and failure correction, as well as a unified training pipeline through which the model learns, within a single framework, capabilities spanning perception, planning with decision-making, and execution with verification.
      \item \textbf{Validation of cross-task generalization.} We show that ME-VLM can be transferred to autonomous driving and embodied robotic tasks, achieving competitive performance in both.
      \item \textbf{Co-optimization for edge deployment.} We perform token compression, quantization, and hardware-software co-optimization targeting edge-side chips, reducing inference latency and computational cost while improving the feasibility of deploying the model in real embodied systems.
\end{itemize}


\section{Related Work}
\label{sec:related-work}

\subsection{Digital Agents}
\label{sec:related-work:digital-agent}
The Qwen series \cite{qwen3_5} have recently continuously strengthened vision–language alignment, long-context processing, and tool invocation. The GLM-5 series \cite{glm5} unify agent, reasoning, and coding capabilities within a single Mixture-of-Experts architecture, and through its asynchronous reinforcement learning infrastructure and Agent RL algorithms substantially improves the model's performance in long-horizon interaction, tool invocation, and self-correction. Kimi K3 \cite{kimi_k3}, built on an MoE architecture with a million-token context window, applies reinforcement learning simultaneously across  long-horizon programming, general-purpose agents, and multimodal reasoning, training the model to complete a full agentic loop over interaction trajectories spanning hundreds to thousands of tool calls.
However, the capabilities of these digital agent models are rooted in the interaction paradigm of the digital world: their tool invocation, reasoning chains, and context management are all oriented toward digital-space tasks such as text, code, and web pages, and lack systematic modeling of spatial structure perception in the physical world, semantic understanding of robot actions, and monitoring of execution states. This limitation is precisely complementary to embodied-domain models, and it is the core motivation for this paper's proposal to unify the two classes of capabilities.

\subsection{Embodied VLMs}
\label{sec:related-work:embodied-vlm}
With the rapid evolution of vision-language models \cite{chen2025mindgpt, qwen2.5, qwen3_5, rynnbrain1_1, hy_embodied_vlm1_0, mimo_embodied, physbrain1.5}, the research community has begun transferring general agent capabilities to physical AI, aiming to build embodied specialists with strong cognitive abilities in the physical world. RynnBrain 1.1 \cite{rynnbrain1_1} strengthens robotic manipulation via contact-point prediction, native 3D spatial perception, and a unified cross-embodiment action space. Hy-Embodied-VLM-1.0 \cite{hy_embodied_vlm1_0} proposes an action-centric three-tier capability taxonomy for physical-world reasoning. Vesta \cite{vesta} unifies localization, navigation, embodied question answering, and long-horizon planning in a single model with multimodal memory. MiMo-Embodied \cite{mimo_embodied} integrates autonomous driving and embodied AI into one VLM, revealing positive cross-domain transfer.  PhysBrain 1.5 \cite{physbrain1.5} unifies environment understanding, action generation, and future-state prediction within a single VLM, trained from human interaction videos.
While these works advance embodied cognition, they offer limited support for multi-turn agent reasoning, skill invocation under open-ended instructions, execution-state monitoring, error diagnosis, and long-horizon planning. To bridge this gap, we jointly train embodied cognition and multimodal agent capabilities as mutually reinforcing objectives within a unified model.


\section{Data Construction}
\label{sec:data-construction}
Digital agents accumulate task-solving experience in the digital world through closed-loop interaction with multimodal context, tools, and skills. Interaction tasks in the physical world, however, demand that these capabilities be grounded in partially observable physical states, spatial constraints, object affordances, and real-world feedback. To this end, our data construction pipeline adopts a unified task-processing paradigm that bridges embodied cognition and interactive capability, building upon digital-world semantic understanding to endow the model with the ability to perceive and understand the real physical world and to autonomously execute physical tasks. 

\subsection{Capability Taxonomy}
\label{sec:data-construction:capability-taxonomy}

\subsubsection{Capability Transfer Rationale}
\label{sec:data-construction:capability-transfer-rationale}
An embodied agent is an extension and dimensional uplift of the digital agent's capabilities, not a capability system built from scratch. Both belong to the same family of agent models: they share a common technical foundation and possess four core underlying capabilities, which also constitute the prerequisites for a digital agent to evolve into an embodied agent: perception of environment, long-horizon task planning, action execution, and diagnosis of real-environment feedback with action correction. Building on this common foundation, the two differ across several dimensions, including scenario boundary, cognitive paradigm, governing rules, and task loop. 

In terms of operating scenarios, digital agents are confined to two-dimensional, deterministic virtual spaces without physical structure or dynamics, whereas embodied agents face three-dimensional, uncertain real physical spaces with complex variables such as physical disturbances. At the perception and cognition level, digital agents possess only digital semantic perception, without 3D spatial modeling, temporal dynamics, or mechanical understanding, whereas embodied agents are equipped with comprehensive physical perception spanning spatial geometry and action execution. In terms of governing rules, digital agents follow digital logic and software conventions, whereas embodied agents must strictly obey objective physical laws, adapting their decisions to physical rules such as force balance and motion constraints. At the task-loop level, the digital agent performs an information-processing loop that iterates over data and information, whereas the embodied agent performs a physical-actuation loop that alters the state of the real physical world through hardware execution.

To accomplish the evolution from digital agent to embodied agent, we define a clear transfer boundary. Capabilities belonging to the general problem-solving structure, including multimodal perception, long-horizon task understanding, tool and skill invocation, and reasoning with attribution, can be effectively transferred from the digital-agent domain to the physical domain \cite{thea,guava}. Capabilities unique to embodied tasks, including spatial state estimation, reasoning about object affordances, modeling of physical action constraints, prediction of action consequences, and closed-loop feedback from the real world, require first establishing cognition of the physical world and cannot be substituted by digital-agent experience.

Our ME-VLM is built precisely on this boundary: while inheriting the general problem-solving capabilities endowed by digital agents, it systematically introduces embodied cognition to address the core weakness of digital agents, their lack of understanding of physical changes, thereby converting general task-planning priors into embodied intelligence that can be reliably executed in real physical environments.

\begin{figure}[!t]
    \centering
    \includegraphics[width=\linewidth]{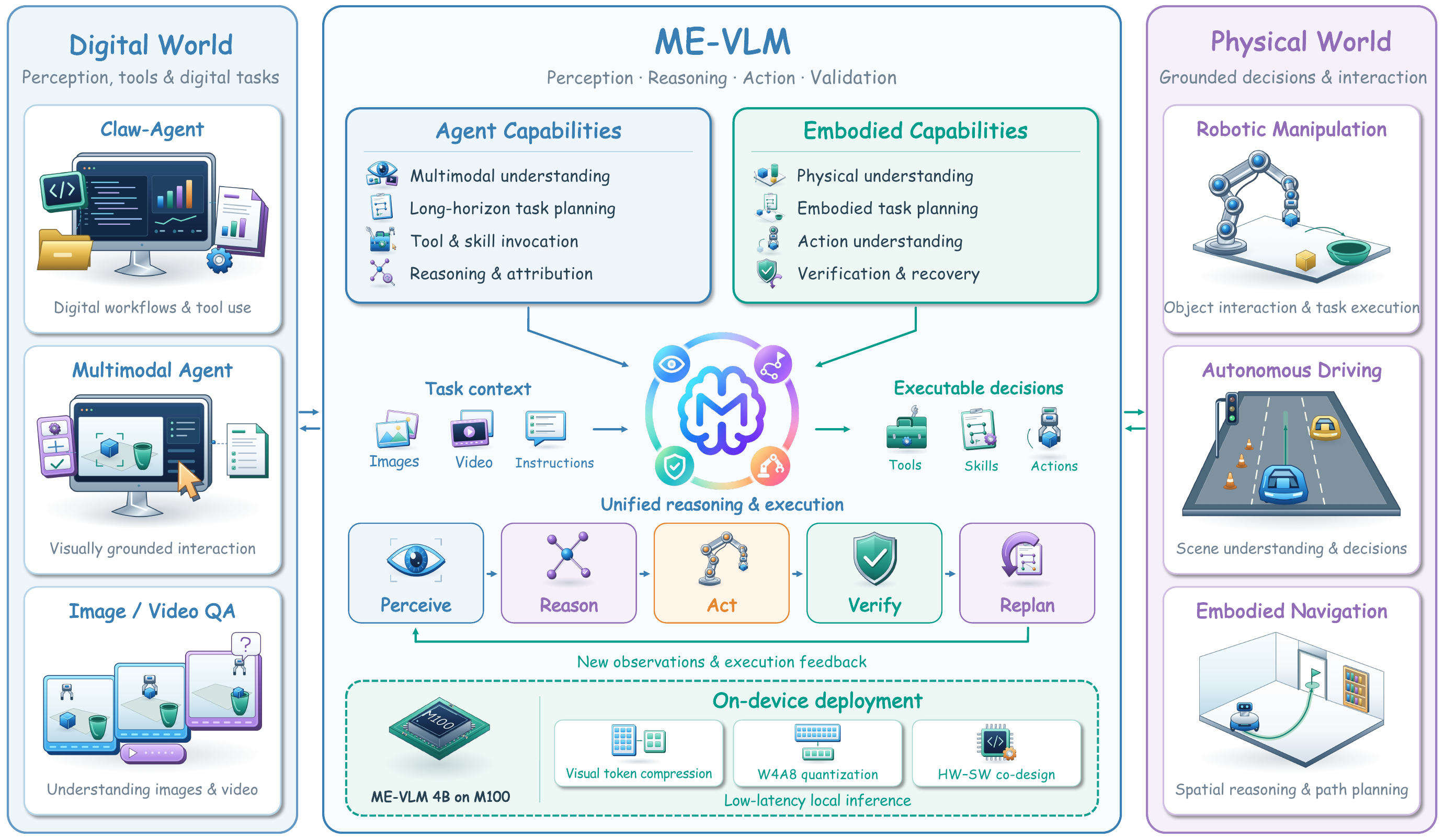}
    \caption{\textbf{Overview of the ME-VLM framework.} ME-VLM integrates agentic and embodied capabilities through a five-stage loop of perception, reasoning, action, verification, and replanning, bridging digital and physical tasks with support for efficient hardware deployment.}
    \label{fig:data:kuangjia}
\end{figure}

\subsubsection{Dual-Capability Cooperation}
\label{sec:data-construction:dual-capability-cooperation}
Embodied intelligence systems are not built by simply stacking physical perception and action-execution modules on top of the capabilities of a digital agent. Rather, they require the construction of a bidirectional driving loop in which embodied cognition and embodied agency are tightly coupled. Within this loop, the embodied agent assumes the global-task function, supplying overall task planning, high-level operational goals, long-horizon contextual association, and the basis for decision-making. Embodied cognition assumes the functions of environmental perception and physical execution, outputting physical-scene states, action-execution results, and environmental feedback signals that underpin cognitive iteration, policy correction, and behavior optimization \cite{roboos,phyagentos}. Together, the two form a collaborative paradigm of ``cognitive feedforward, physical feedback, dynamic correction, and continuous iteration.'' This layered collaboration mechanism aligns with the dual cognitive–motor system logic of the human brain: the prefrontal cortex is responsible for goal setting, long-range planning, and action monitoring and diagnosis, corresponding to agency capability; the sensorimotor cortex is responsible for scene perception, limb regulation, and the grounded execution of actions, corresponding to embodied physical cognition capability. Concretely, cognitive capability and agency capability construct a complete embodied-intelligence closed-loop system through four core collaborative relationships:

\textbf{First, multimodal perception evolves into understanding of physical state.} The multimodal perception of an agent remains at the level of recognizing image, text, and interface semantics, with the core objective of parsing digital information. It lacks awareness of the physical attributes. In the bidirectional closed-loop system of an embodied agent, multimodal perception is upgraded from mere semantic-information recognition to element- and space-level perception oriented toward the physical scene. Leveraging cross-modal alignment capability, the model not only recognizes object categories and scene semantics but further perceives object spatial position, geometric structure, relative relationships, and dynamic-change trends, achieving a fine-grained understanding of the global state of the physical scene \cite{rynnbrain1_1}.

\textbf{Second, long-horizon task understanding constrains embodied global planning.} The long-horizon reasoning and global context-modeling capabilities of the agent provide cross-subtask, cross-step overall goal constraints for embodied task planning, so that every decision made during physical execution is subordinated to the global task logic \cite{vesta,roboos}. This prevents the localized bias of single-step optimality at the cost of global mismatch, and realizes structured, orderly, long-horizon planning of physical operations.

\textbf{Third, tool and skill invocation empowers embodied action execution.} The tool and skill capabilities of the agent take on the function of translating semantics from the cognitive to the physical-execution ability, converting high-level task-planning intent into structured semantic instructions recognizable for robot action understanding, trajectory planning, grasping execution, and navigational movement. Here, external tools supplement environmental and external information for physical reasoning, while internal tools encapsulate foundational robotic capabilities such as object detection and grasp planning into standardized semantic-skill interfaces \cite{guava,thea}. Through the Agentic Loop, the agent orchestrates and invokes these on demand, performing adaptive matching in accordance with task characteristics, observation modality, and action semantics, and thereby achieves modular, compositional execution of complex physical tasks. This constitutes the core bridge through which cognitive intent is grounded as physical action.

\textbf{Fourth, reasoning and attribution capability support verification and error recovery.} The causal reasoning and task attribution of the agent provide the core decision basis for anomaly checking, error correction, and fault recovery in the embodied system. Embodied cognition can only perceive execution anomalies and state deviations. It cannot distinguish whether the root cause of an error originates from perception noise, planning bias, or environmental disturbance. The embodied-agent capability can conduct a global postmortem in combination with task instructions, the trajectory of environmental state evolution, and historical interaction experience, pinpoint the cause of an error, and generate a corrective strategy \cite{embodiskill,thea}. The stronger the attribution and reasoning capability, the higher the system's learning efficiency from failure samples, and the higher the upper bound of the self-evolutionary iteration of physical AI.

These four mechanisms form a bidirectional closed loop: agency capability drives embodied execution while physical feedback nourishes cognitive evolution, delineating a plausible path from a digital agent to a physical agent. As shown in Figure~\ref{fig:data:kuangjia}, this loop is operationalized as a closed perception–reasoning–action–verification–replanning cycle that integrates agentic and embodied capabilities, re-grounding digital-world experience for physical-world perception and autonomous execution.

\begin{figure}[!t]
    \centering
    \includegraphics[width=\linewidth]{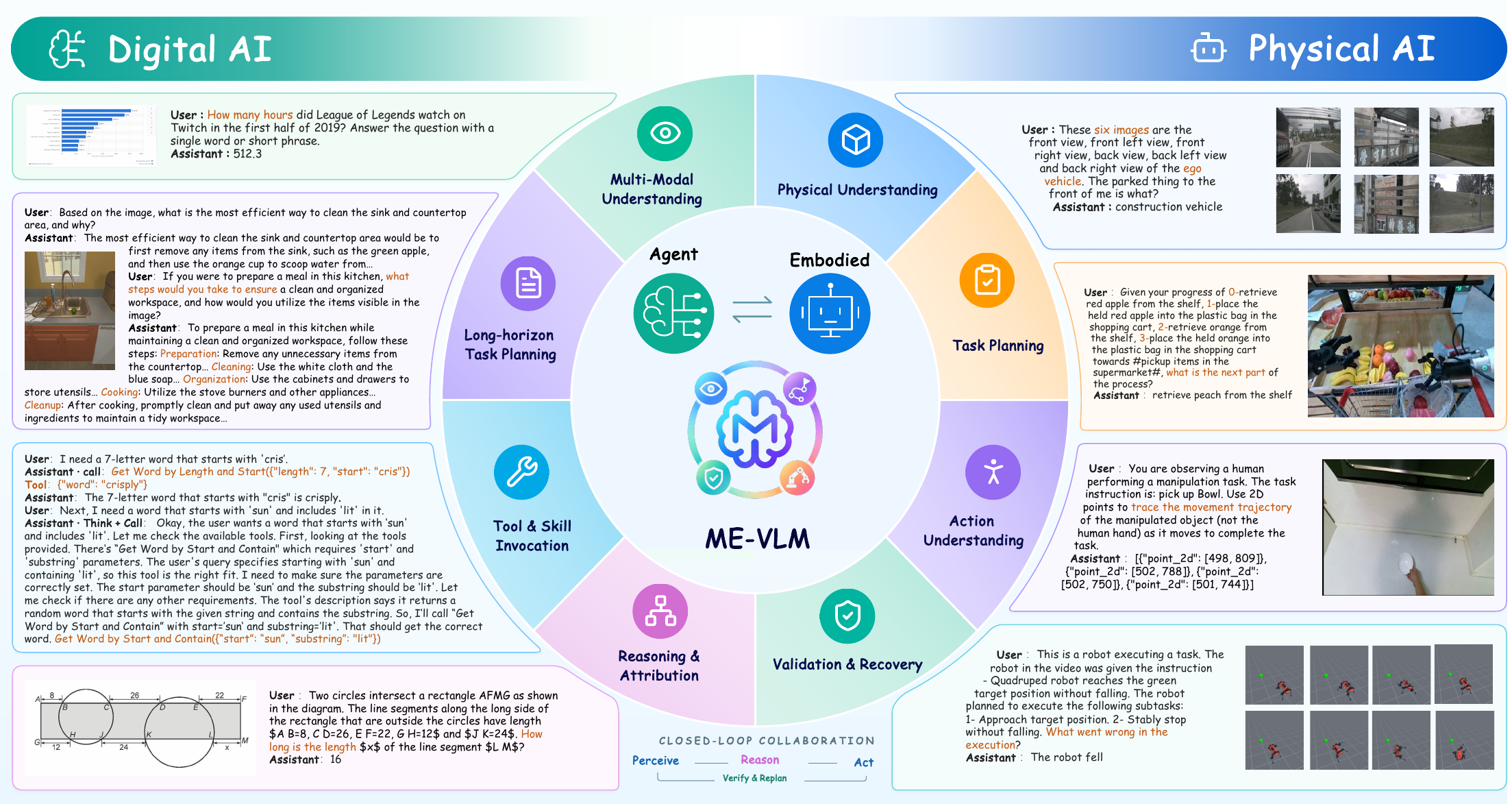}
    \caption{\textbf{Representative data examples for ME-VLM.} Digital AI data cover multimodal understanding, long-horizon task planning, tool and skill invocation, and reasoning and attribution. Physical AI data cover physical understanding, task planning, action understanding, and validation and recovery.}
    \label{fig:data:data_all}
\end{figure}

\begin{figure*}[t]
    \centering
    \includegraphics[width=\textwidth]{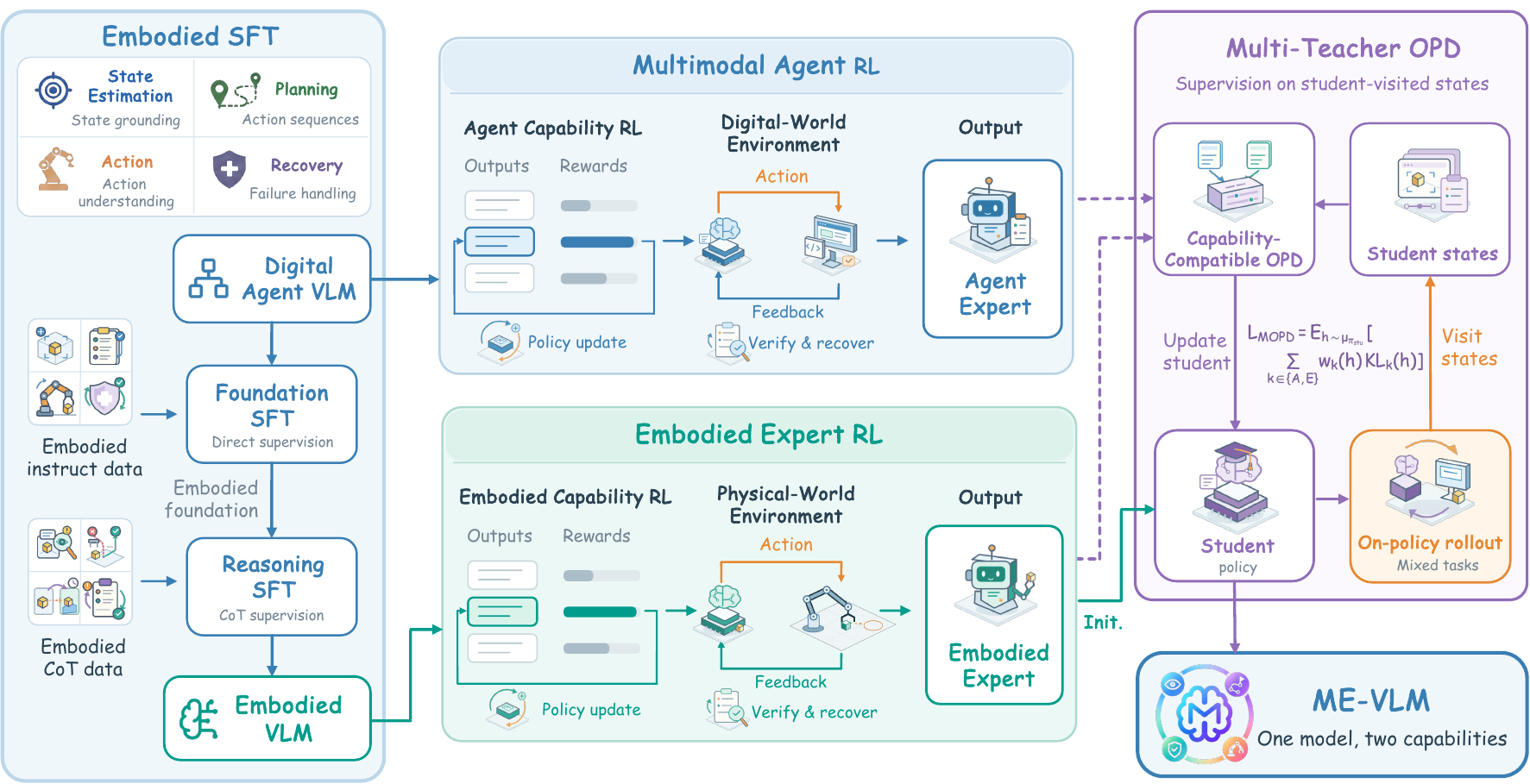}
    \caption{\textbf{Overview of the ME-VLM training strategy.} The original
    base policy initializes multimodal-agent RL, while the embodied SFT
    checkpoint initializes embodied-expert RL. Each branch combines
    capability-unit optimization with sandboxed environment-loop training.
    The resulting experts supervise student-visited states and are distilled
    into a single deployable model.}
    \label{fig:training-strategy}
\end{figure*}

\subsection{Data Curation}
\label{sec:data-construction:production-curation}
As illustrated in Figure~\ref{fig:data:data_all}, we construct capability-system data following the aforementioned construction paradigm for embodied agent. The dataset includes capability data for both digital and physical agents. We elaborate on the construction methodology of these capability data in the following subsection.

\subsubsection{Embodied Data}
\label{sec:data-construction:embodied-cognitive-data}
Data for embodied capabilities is drawn from both open-source and self-produced sources. The open-source portion comprises four categories of capability data: physical understanding, embodied task planning, error diagnosis and recovery. In addition, we construct recovery data based on virtual embodied environments. Specifically, action models \cite{black2025pi05, lingbot_vla, lingbotvla2} first execute designated tasks, during which offset perturbations are injected at random moments and the resulting anomalous states are saved; the model then resumes execution from each perturbed state. After validity checking, trajectories are filtered by task, perturbation parameters, and outcome, retaining at most one successful and one failed trajectory per perturbed state. The complete trajectories are then preserved per method, labeled, and converted into the training format, with state isolation between training and evaluation enforced throughout.

Autonomous driving data come from open-source datasets and cover four capability categories: panoramic scene understanding, long-tail traffic scene perception, risky object recognition, and perception-prediction-planning closed-loop decision-making. Embodied navigation data is produced by our in-house data pipeline: we build training data on two underlying navigation datasets (R2R \cite{r2r} and RxR \cite{rxr} scenes). The simulator collects first-person observations and atomic actions along ground-truth paths; consecutive actions in the same direction are merged into high-level actions, and the data is uniformly reorganized into a multi-turn dialogue format consistent with the evaluator. The system prompt declares a closed action space (forward 25/50/75 cm; turning 15/30/45° for R2R and 30/60/90° for RxR), each turn introduces only a single new observation frame. Each response is a bare action list; a single STOP action terminates the episode. For each source, two decision granularities are provided: one action per turn and up to three actions per turn.

\subsubsection{Agent Data}
\label{sec:data-construction:multimodal-agentic-data}
Agent data comprise open-source tool-use tasks and self-constructed multimodal tasks. The open-source portion provides user requests and tool specifications, organized into a consistent task format. The self-constructed portion combines visual or environmental context with user goals, task constraints, available tools, and subtask dependencies, emphasizing tasks that require visual understanding, multi-step planning, and tool coordination. Each task is paired with evaluation criteria for assessing response correctness or task completion. Interactive tasks additionally specify an initial environment state and executable tool interfaces, allowing the model to observe action outcomes and adjust subsequent decisions. Together, these tasks support the development of perception, planning, tool use, and long-horizon state tracking in complex scenarios.


\section{Training Strategy}
\label{sec:training-strategy}

ME-VLM is built on a Qwen3.5-series native multimodal agent model
\cite{qwen3_5}. We denote the original starting policy by
$\pi_{\theta_0}$. Our training transfers its multimodal understanding,
planning, tool-use priors and complex reasoning into decisions grounded in physical observations and environment feedback. This transfer follows the VLA paradigm of adapting
large-scale vision-language representations to physical decision making
\cite{brohan2023rt2,kim2024openvla,black2025pi05}.

As illustrated in Figure~\ref{fig:training-strategy}, realizing this transfer
requires resolving three successive optimization challenges: grounding general
semantic priors in physical states and actions, improving agentic and embodied
policies without destructive interference, and consolidating their
complementary strengths into a single deployable model. The following sections
address these challenges in this order.

\subsection{Embodied Capability Injection}
\label{sec:embodied-capability-injection}

Multimodal agent models provide useful semantic and planning priors,
but those priors do not by themselves ensure that a proposed action is
feasible for a particular embodiment or valid in the current physical state
\cite{ahn2022saycan,zawalski2024ecot}. We therefore adapt the base multimodal agent through two successive SFT stages that connect pretrained semantic representations to
continuous observations, physical states, and actions.

\noindent\textbf{Embodied Foundation SFT.}
This stage establishes broad embodied competence and stable physical
grounding. Using the directly supervised tasks described in
Section~\ref{sec:data-construction}, the
model learns to associate multimodal observations with action-relevant
physical states, including spatial relations, object affordances, action
preconditions, feasibility, and expected consequences. Its purpose is to
align the semantic representations of the base model with action-relevant
physical concepts, thereby establishing transferable embodied representations
and reliable observation-to-decision mappings for subsequent reasoning,
planning, and policy optimization. Recent VLA systems similarly benefit from
co-training semantic predictions, visual observations, and physical actions
\cite{kim2024openvla,black2025pi05}.

\noindent\textbf{Embodied Reasoning SFT.}
Starting from the foundation checkpoint, this stage trains the model to
apply its acquired physical knowledge to embodied decision-making.
Given the current observation and task state, the model is supervised
to identify the conditions relevant to the task, assess possible actions
under physical constraints, and select an appropriate action. It also
learns to predict the resulting state change and specify how to verify
whether the action achieves its intended effect. Together, these steps
ground reasoning in the visual entities, robot states, plans, and motions
involved in task execution, consistent with the role of an embodied
brain \cite{zawalski2024ecot}. The emphasis is on using this supervision
to develop representations that support physically grounded decisions,
rather than requiring the model to generate detailed reasoning at
inference time, as suggested by recent findings
\cite{sun2026revisitingecot}. The resulting policy
$\pi_{\theta_{\mathrm{inj}}}$ provides a physically grounded and
reasoning-aware initialization for the embodied expert RL branch.

\subsection{Expert Reinforcement Learning}
\label{sec:expert-rl}

We train the two experts from different initializations. The multimodal agent
expert $\pi_{\theta_{\mathrm{A}}}$ is initialized directly from the original
base policy $\pi_{\theta_0}$, preserving its general multimodal
reasoning, planning, and tool-use capabilities. The embodied cognition expert
$\pi_{\theta_{\mathrm{E}}}$ is initialized from
$\pi_{\theta_{\mathrm{inj}}}$, inheriting the physical grounding and embodied
decision structure acquired through SFT. This asymmetric initialization
reflects the different optimization priorities of the two domains:
agent tasks emphasize goal decomposition, tool orchestration, and long-horizon
state tracking, whereas embodied tasks emphasize spatial accuracy, physical
feasibility, and state-transition consistency. Directly mixing heterogeneous
RL objectives can weaken domain-specific capabilities, whereas independently
trained teachers avoid unwanted coupling between the two optimization
objectives \cite{ma2026mopd}. We
therefore specialize the experts first and reconcile
them only after domain-specific optimization.

\subsubsection{Capability-Unit RL}
\label{sec:capability-unit-rl}

Capability-unit RL improves verifiable next-step decisions, including
reasoning results, spatial predictions, subgoal selection, tool calls, action
feasibility judgments, and local embodied actions. For an input $\mathbf{x}$
of task type $\tau$, we sample $G$ responses
$\{\mathbf{y}_i\}_{i=1}^{G}$ from the old expert policy and evaluate each with
a task-compatible reward $r_i=R_{\tau}(\mathbf{x},\mathbf{y}_i)$. Rewards are
calibrated within each task family before heterogeneous tasks are mixed. We
use the group-relative reward normalization introduced by GRPO
\cite{shao2024deepseekmath}, together with the sequence-level importance ratio
and clipping of Group Sequence Policy Optimization (GSPO)
\cite{zheng2025gspo}:
\begin{align}
    \widehat{A}_i
    &= \frac{r_i-\mu_G}{\sigma_G+\epsilon},
    \quad
    \mu_G=\frac{1}{G}\sum_{i=1}^{G}r_i,
    \label{eq:group-relative-advantage}\\
    s_i^{k}(\theta_k)
    &= \left(
        \frac{
            \pi_{\theta_k}(\mathbf{y}_i\mid\mathbf{x})
        }{
            \pi_{\theta_k^{\mathrm{old}}}(\mathbf{y}_i\mid\mathbf{x})
        }
    \right)^{\!1/T_i}
    = \exp\!\left[
        \frac{1}{T_i}\sum_{j=1}^{T_i}
        \log
        \frac{
            \pi_{\theta_k}(y_{i,j}\mid\mathbf{x},\mathbf{y}_{i,<j})
        }{
            \pi_{\theta_k^{\mathrm{old}}}
            (y_{i,j}\mid\mathbf{x},\mathbf{y}_{i,<j})
        }
    \right],
    \label{eq:gspo-sequence-ratio}\\
    \mathcal{L}_{\mathrm{CU}}^{k}
    &= -\mathbb{E}\!\left[
        \frac{1}{G}\sum_{i=1}^{G}
        \min\!\left(
            s_i^{k}(\theta_k)\widehat{A}_i,
            \operatorname{clip}
            (s_i^{k}(\theta_k),1-\varepsilon,1+\varepsilon)
            \widehat{A}_i
        \right)
    \right],
    \label{eq:capability-unit-objective}
\end{align}
where $k\in\{\mathrm{A},\mathrm{E}\}$ and $T_i=|\mathbf{y}_i|$. The
length-normalized ratio $s_i^{k}$ assigns a single importance weight to the
entire response, aligning off-policy correction and clipping with the
sequence-level reward. The reward follows the structure of the prediction:
exact or format-aware rewards are used for discrete outputs, geometric metrics
for spatial outputs, and partial-credit or verifier-based rewards for plans
and open-ended decisions. This avoids forcing heterogeneous capabilities into
a single binary correctness signal.

\subsubsection{Environment-Loop RL}
\label{sec:environment-loop-rl}

Environment-Loop RL trains the experts through executable multi-turn
interaction rather than isolated response evaluation. Each rollout runs in an
independent, task-specific sandbox or simulator instance initialized from a
reproducible snapshot. The initialization specifies the starting environment
state together with the task-scoped tools, Skills, resources, and software
dependencies required for execution. Rollouts sampled for the same task are
initialized from equivalent snapshots, making their outcomes comparable
within a GSPO group. After a rollout terminates, its environment is discarded
or reset, preventing side effects and state changes from leaking into other
trajectories. Executable gyms and sandboxed rollout services have become core
infrastructure for recent agent-RL systems
\cite{jain2025r2egym,zhang2026prorlagent}.

Within each sandbox, training follows an explicit interaction loop. The model
receives the current observation and interaction history, generates a tool or
skill call, and submits it to the environment executor. The sandbox validates
and executes the action, records its side effects, and returns the resulting
observation, execution status, or error message. This feedback is appended to
the context for the next decision, and the loop continues until the task
succeeds, reaches a failure condition, or exceeds its interaction budget.
Execution failures are retained as part of the trajectory so that the model
can learn verification, correction, and recovery rather than being trained
only on successful paths. Recent multi-turn agent-RL frameworks similarly
emphasize stochastic environment feedback, execution--training separation, and
explicit trajectory decomposition \cite{wang2025ragen,luo2025agentlightning}.

Evaluation is coupled to the environment instead of relying on a generic
response-level judge. In digital-agent sandboxes, task-specific evaluators
inspect tool outputs and the resulting page, file, resource, or artifact
state. In embodied environments, evaluators inspect simulator states, goal
predicates, physical state transitions, and constraint violations. These
environment-grounded signals combine terminal task success with verifiable
intermediate events, enabling finer-grained credit assignment across the
multi-turn loop \cite{wei2026multiturn}. This stage produces the two
teacher policies used in the final capability-reconciliation stage.

\subsection{Multi-Teacher On-Policy Distillation}
\label{sec:mopd}

The two experts are not retained as an inference-time ensemble. Instead, we
distill them into a deployment-scale student initialized from the embodied
expert, $\theta_{\mathrm{stu}}^{(0)}\leftarrow\theta_{\mathrm{E}}$. This
\emph{embodied anchoring} preserves the physically grounded representation as
the starting point, while the multimodal agent teacher restores and extends
general planning and tool-use capabilities. This follows the MOPD paradigm of
first obtaining domain-specialized RL teachers and then consolidating them in
a single student on its own rollout distribution \cite{ma2026mopd}, without
introducing expert routing at inference time.

Offline distillation exposes the student primarily to teacher-generated
trajectories and therefore does not directly correct errors in states reached
by the student itself. On-policy distillation addresses this mismatch by
querying teachers on student-generated prefixes
\cite{agarwal2024gkd,ma2026mopd}. Recent analyses further show that successful
OPD depends on compatible teacher--student behavior and genuinely novel
teacher capabilities \cite{li2026rethinkingopd}. During
MOPD, the student rolls out on a mixture of agent and embodied tasks. At a
student prefix
$\mathbf{h}^{\mathrm{stu}}_t=(\mathbf{x},\mathbf{y}^{\mathrm{stu}}_{<t})$,
the valid teachers provide distributions over the same token or structured
action space. We minimize the reverse KL divergence from the student to each
valid teacher:
\begin{equation}
    \mathcal{L}_{\mathrm{MOPD}}
    = \mathbb{E}_{
        \mathbf{h}^{\mathrm{stu}}_t
        \sim\mu_{\pi_{\theta_{\mathrm{stu}}}}
    }
    \left[
        \sum_{k\in\mathcal{K}_t}
        w_k(\mathbf{h}^{\mathrm{stu}}_t)
        D_{\mathrm{KL}}\!\left(
            \pi_{\theta_{\mathrm{stu}}}
            (\cdot\mid\mathbf{h}^{\mathrm{stu}}_t)
            \,\middle\|\,
            \pi_{\theta_k}(\cdot\mid\mathbf{h}^{\mathrm{stu}}_t)
        \right)
    \right],
    \label{eq:mopd-objective}
\end{equation}
where $\mathcal{K}_t\subseteq\{\mathrm{A},\mathrm{E}\}$ contains the teachers
valid for the current decision, and
$w_k\geq0$ with $\sum_{k\in\mathcal{K}_t}w_k=1$. Agent decisions assign more
weight to $\pi_{\theta_{\mathrm{A}}}$, while embodied decisions assign more
weight to $\pi_{\theta_{\mathrm{E}}}$. These weights are used only during
training. If the models do not share a tokenizer or action encoding,
distillation is performed after mapping their outputs to a common structured
representation.

MOPD thus transfers complementary expert behavior on the state distribution
that the final student is likely to encounter. The result is a single policy
that preserves physical grounding while supporting general multimodal
planning, tool use, execution verification, and recovery.

\section{Edge Chip Deployment}
\label{sec:edge-chip-deployment}

\begin{figure}[!t]
    \centering
    \includegraphics[width=\linewidth]{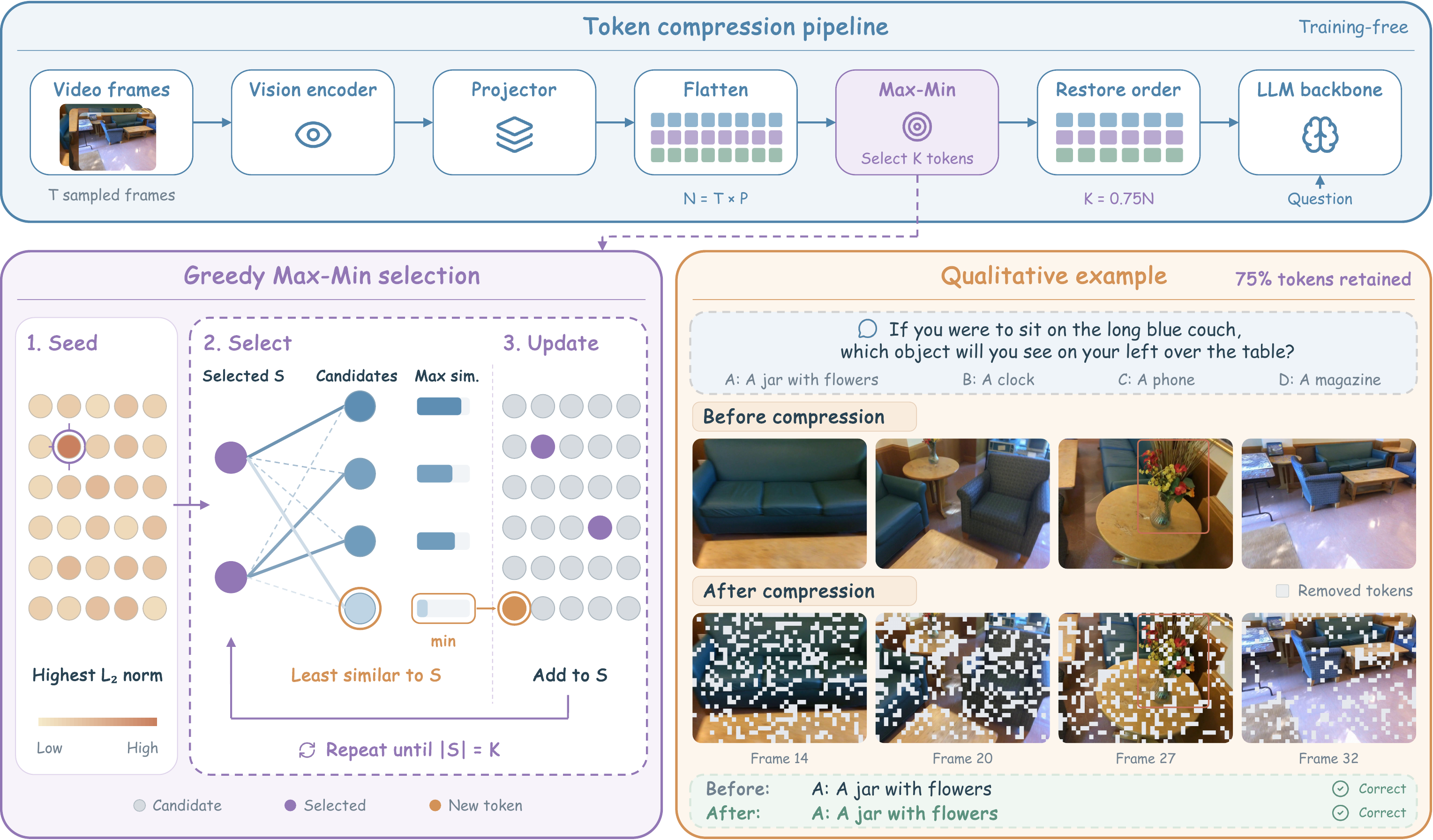}
    \caption{\textbf{Training-free visual token compression via greedy Max--Min selection.} The figure illustrates the compression pipeline, diversity-based token selection, and a qualitative example with 75\% of visual tokens retained.}
    \label{fig:edge-token-compression}
\end{figure}

\begin{figure}[t]
    \centering
    \includegraphics[width=\linewidth]{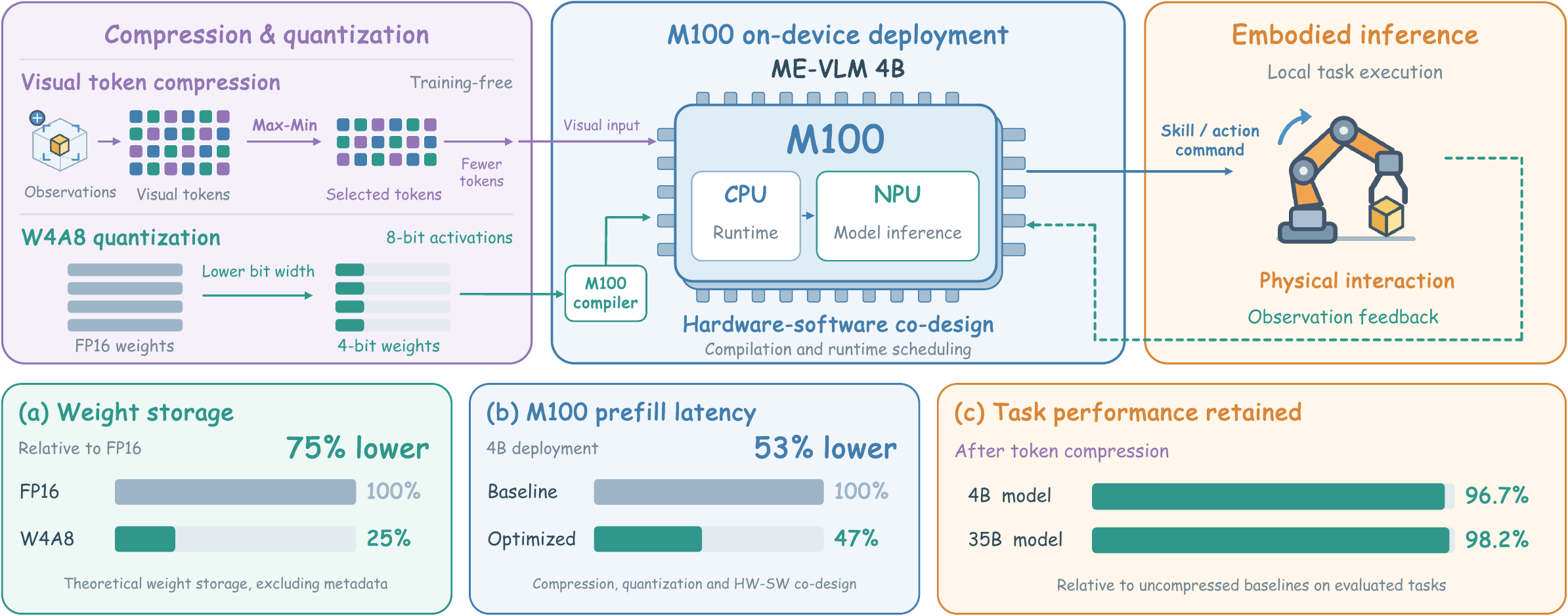}
    \caption{\textbf{ME-VLM deployment on M100} with visual token compression, W4A8 quantization, and hardware--software co-design. Results summarize storage savings, prefill acceleration, and task-performance retention.}
    \label{fig:m100-deployment}
\end{figure}

Embodied tasks involving interaction with the physical world impose stringent requirements on inference latency, adaptability to complex environments, data security, and operational autonomy. Local inference on edge hardware is therefore essential for practical deployment at scale. However, deploying large embodied foundation models on edge devices introduces several system-level challenges, including limited compute capacity, power constraints, hardware architecture mismatches, and difficulties in software adaptation. These constraints create a three-way trade-off among inference accuracy, response latency, and power consumption.

We address these challenges through complementary optimizations at the model and system levels. At the model level, visual token compression reduces the computational burden of the inference pipeline by compacting visual inputs before they enter the language model. At the system level, hardware--software co-design adapts the inference execution framework, sensor synchronization and scheduling mechanisms, and hardware acceleration interfaces to embodied workloads. Together, these optimizations establish an integrated execution and scheduling path spanning perception, physical reasoning, and motion planning and control. By jointly adapting algorithms and hardware, we bridge the gap between AI software developed for digital environments and embodied tasks in the physical world, enabling deployment at scale on edge devices.

\subsection{Edge Acceleration Strategy}
\label{sec:edge-chip-deployment:acceleration-strategy}

The inputs to embodied policy models consist primarily of visual observations. After processing by the vision encoder and projector, individual frames, multi-view images, or videos produce a large number of visual tokens. These tokens substantially outnumber text tokens and account for most of the prefill computation on edge devices. Visual inputs in embodied scenarios also contain considerable redundancy~\cite{tao2025dycoke}: static backgrounds across consecutive frames and overlapping regions across adjacent views cause near-duplicate features to repeatedly consume the limited token budget. We therefore introduce visual token compression to reduce the number of tokens passed to the language model while preserving the model's capabilities.

For practical deployment, we adopt a training-free, plug-and-play token selection scheme. As illustrated in Figure~\ref{fig:edge-token-compression}, the compression module is inserted between the visual projector and the large language model backbone, where it selects a subset of visual tokens that have already been mapped into the language model's embedding space. The selected visual tokens are then fed into the language model together with the text tokens. The architecture and parameters of the vision encoder, projector, and language model remain unchanged. This approach requires no additional training and can be applied directly to checkpoints from any training stage. Selection only removes tokens; it neither merges tokens nor modifies their features, preserving the feature representations of the retained tokens as learned during training. The retained token count, $K$, is a tunable parameter configured independently of the model according to the compute and latency budgets of the target device.

Let the visual token set produced by the projector be denoted by $X = \{x_1, \ldots, x_N\}$, where $N$ is the product of the number of frames or views and the number of tokens per frame or view. The compression module retains $K < N$ tokens from $X$. Rather than assigning an importance score to each token independently~\cite{chen2025fastv}, we formulate selection as a diversity maximization problem in feature space~\cite{ravi1994dispersion}. Specifically, we select the subset of $K$ tokens that maximizes the minimum pairwise distance:
\begin{equation}
    S^{*} = \operatorname*{arg\,max}_{\substack{S \subset X \\ |S| = K}} \; \min_{\substack{x_i, x_j \in S \\ i \neq j}} d(x_i, x_j),
    \label{eq:edge-maxmin-selection}
\end{equation}
where $d(x_i, x_j)$ denotes the cosine distance between two visual token representations.

This objective prevents near-duplicate features from dominating the token budget, allowing the retained set to cover a broad range of visual content. Since exact optimization is computationally expensive, we approximate the solution using a greedy Max-Min algorithm~\cite{alvar2025divprune}. We initialize the retained set with the token having the largest $\ell_2$ norm. At each subsequent iteration, we select the candidate whose maximum similarity to any token in the retained set is smallest, continuing until $K$ tokens have been retained. Each iteration thus prioritizes visual features that are not yet represented in the selected set, preserving diversity while removing redundancy. Once selection is complete, the retained tokens are reordered according to their original sequence indices to preserve their relative spatiotemporal structure. No token features are modified during this process.

For multi-view image or multi-stream video inputs, we apply the selection procedure independently within each stream using the same retention ratio, then organize the retained tokens according to the original input structure. This prevents the diversity objective from favoring a single view and accommodates visual inputs with varying numbers of views, resolutions, and sequence lengths.

\subsection{Co-Design for M100}
\label{sec:edge-chip-deployment:mh100}

We deploy the 4B model on Li Auto's M100 system-on-chip (SoC)~\cite{xie2026m100} for local inference, enabling embodied systems to operate in real-world environments without relying on remote GPU servers. The M100 SoC integrates an in-house M100 neural processing unit (NPU) designed for general-purpose AI inference, a 24-core Arm CPU subsystem, and an LPDDR5X memory subsystem supporting up to 64~GB of memory and a peak bandwidth of 273~GB/s. The Arm CPU handles application logic and inference runtime management, while the M100 NPU executes most tensor computations required by the embodied policy model. The NPU uses a compiler-orchestrated dataflow architecture, in which computation and data exchange across compute units are coordinated through software-controlled data movement between on-chip SRAM and external LPDDR5X memory. Pipelined overlap between computation and data transfers improves overall execution efficiency.

To further reduce memory requirements and memory access overhead during edge inference, we adopt W4A8 quantization: model weights are quantized from FP16 to 4~bits, while inference activations use 8~bits. Compared with FP16 weights, 4-bit weights theoretically reduce weight storage by 75\%, substantially decreasing the model's memory footprint and the volume of data transferred when loading weights from external memory. The actual model footprint is slightly larger than this theoretical minimum because quantization scale factors and other metadata also require storage. Ablation experiments on embodied task benchmarks show less than 1\% performance degradation after quantization, demonstrating that W4A8 reduces inference overhead without materially compromising model performance. Figure~\ref{fig:m100-deployment} provides an overview of the M100 deployment workflow and summarizes the effects of quantization, joint prefill optimization, and visual token compression.

\subsubsection{Joint HW-SW Optimization for Prefill}
\label{sec:edge-chip-deployment:prefill-optimization}

The prefill stage processes long input token sequences and accounts for a substantial share of inference latency on edge devices. Feeding the full token sequence into M100 exposes substantial computational parallelism but also incurs a large computational workload. Conversely, excessive token compression can leave too little work to fully utilize the NPU's parallel compute capacity. We therefore jointly optimize model-level token compression and execution on M100 to identify an operating point that balances task accuracy, total computation, and hardware utilization.

Specifically, we use W4A8 quantization together with token compression applied to the base model. Token compression reduces the number of tokens processed during prefill and subsequent computation from 4,096 to 3,328. We select this token budget based on the M100 NPU's parallel execution capacity and profiling of real application workloads, subject to task accuracy requirements. This ensures that the resulting workload is well matched to the NPU resources allocated to the embodied model. Token compression directly reduces prefill computation, while a sufficiently large retained token budget avoids underutilization caused by excessive compression and maintains high NPU utilization for the core tensor operations. In deployment measurements, this joint optimization reduces prefill latency from 400~ms to 188~ms, a reduction of approximately 53\% and a prefill speedup of approximately $2.13\times$. We further evaluate the impact of visual token compression on task performance. On the evaluated tasks, the 4B and 35B models retain 96.7\% and 98.2\% of their respective uncompressed baseline scores after token selection. These results show that the token selection strategy reduces the visual input size while maintaining task performance close to the uncompressed baselines, supporting more efficient inference on edge devices.

\subsubsection{Compilation \& Runtime Pipeline}
\label{sec:edge-chip-deployment:compilation-runtime}

The trained embodied model is compiled into an executable binary for edge deployment using the M100 compilation toolchain~\cite{xie2026m100}. The compiler performs computational graph optimization, tensor tiling, and scheduling, mapping tensor operations, data transfers, and synchronization primitives onto M100 NPU resources. For key operators in the model, the compiler exploits spatial parallelism across compute blocks and opportunities for pipelined execution to improve overall hardware utilization.

The runtime software consists of two layers: an AI inference runtime executing on the SoC's Arm CPU cores and NPU firmware executing on the RISC-V processor within the M100 NPU. The inference runtime manages model loading, input preprocessing, inference task submission, and output postprocessing. The NPU firmware handles low-level tensor instruction dispatch, compute-unit scheduling, and data transfers between on-chip and external memory. Together, these components provide a complete on-device inference pipeline from sensor acquisition to robot policy output.

\begin{table*}[!t]
\caption{\textbf{Results on embodied benchmarks.}}
\label{tab:embodied_benchmarks}
\centering
\renewcommand{\arraystretch}{1.28}
\setlength{\tabcolsep}{2.6pt}
\setlength{\arrayrulewidth}{0.55pt}
\arrayrulecolor{black}
\normalsize
\begin{adjustbox}{max width=\textwidth}
\begin{tabular}{@{}ll|>{\columncolor{LiLight}}c>{\columncolor{LiLight}}c|cccc@{}}
\hline
\multirow{2}{*}{\textbf{Subcategory}} & \multirow{2}{*}{\textbf{Benchmark}} & ME-VLM & ME-VLM & Hy-Embodied & RynnBrain 1.1 & PhysBrain 1.5 & MiMo-Embodied \\
& & 4B & 35B-A3B & 30B-A3B & 9B & 8B & 7B \\
\hline
\multicolumn{2}{l|}{Average} & 63.4 & 70.9 & 60.2 & 62.1 & 62.7 & 54.3 \\
\hline
\multirow{15}{*}{\shortstack[l]{Physical \\ Understanding}}
& CV-Bench & 89.5 & 91.7 & 89.7 & 88.2$^*$ & 90.0 & 88.8 \\
& VABench-Point & 67.0 & 75.4 & 59.7 & 20.3$^*$ & 65.2 & 33.3$^*$ \\
& VABench-Visual & 81.6 & 82.6 & 79.7 & 87.6$^*$ & 89.8 & 66.9$^*$ \\
& RefCOCO-testA & 85.9 & 87.8 & 49.4$^*$ & 89.5$^*$ & 60.9$^*$ & 74.5$^*$ \\
& RefCOCO-testB & 76.6 & 80.3 & 56.2$^*$ & 81.6$^*$ & 62.8$^*$ & 67.4$^*$ \\
& ERQA & 45.8 & 54.3 & 60.8 & 47.5 & 52.8 & 46.8 \\
& MindCube & $-$ & 91.1 & 70.0 & 86.9 & 86.2 & 35.2$^*$ \\
& SPARBench & 57.3 & 67.6 & 53.4$^*$ & 51.1$^*$ & 53.8$^*$ & 41.2$^*$ \\
& SPBench-MV & 69.9 & 83.5 & 59.8$^*$ & 74.4$^*$ & 78.8$^*$ & 49.4$^*$ \\
& SPBench-SI & 70.6 & 71.7 & 50.5$^*$ & 59.0$^*$ & 77.7$^*$ & 44.8$^*$ \\
& MMSI-Bench & 33.3 & 46.4 & 41.8 & 47.0 & 41.0 & 29.6 \\
& ViewSpatial-Bench & 57.8 & 62.7 & 53.3 & 54.2$^*$ & 62.5 & 40.5$^*$ \\
& EmbSpatial-Bench & 81.0 & 83.1 & 82.7 & 81.9 & 81.8 & 76.2 \\
& RefSpatial-Bench & 46.9 & 63.4 & 53.4 & 67.2 & 50.9 & 48.0 \\
& RoboSpatial-Home & 67.7 & 74.0 & 69.4 & 69.1 & 73.9 & 61.8 \\
\hline
\multirow{3}{*}{\shortstack[l]{Task \\ Planning}}
& Cosmos & 65.7 & 76.9 & 66.9 & 56.1$^*$ & 72.8 & 56.8 \\
& EgoPlan2 & 55.3 & 64.5 & 49.6 & 43.5$^*$ & 62.1 & 43.0 \\
& RoboBench-Planning & 40.1 & 53.7 & 54.9 & 59.8$^*$ & 42.2$^*$ & 57.2$^*$ \\
\hline
\multirow{4}{*}{\shortstack[l]{Action \\ Execution}}
& VSIBench & 63.2 & 71.2 & 58.9 & 74.9 & 61.9 & 48.5 \\
& SITE-Bench-Video & 65.6 & 72.9 & 69.2 & 68.9$^*$ & 67.5$^*$ & 59.0$^*$ \\
& RoboBench-Perception & 41.8 & 52.2 & 55.9$^*$ & 40.5$^*$ & 31.8$^*$ & 34.9$^*$ \\
& RoboBench-Affordance & 47.0 & 60.7 & 61.7$^*$ & 33.0$^*$ & 25.6$^*$ & 36.7$^*$ \\
\hline
\multirow{4}{*}{Correction}
& RoboFail-Execution & 79.7 & 86.3 & 62.8$^*$ & 77.1$^*$ & 77.8$^*$ & 72.6$^*$ \\
& RoboFail-Planning & 56.7 & 63.3 & 53.3$^*$ & 63.3$^*$ & 56.7$^*$ & 63.3$^*$ \\
& RoboFAC & 74.7 & 75.7 & 51.0 & 55.4$^*$ & 64.8$^*$ & 61.2$^*$ \\
& RoboBench-Error & $-$ & 52.7 & 53.0$^*$ & 42.2$^*$ & 34.1$^*$ & 33.5$^*$ \\
\hline
\end{tabular}
\end{adjustbox}

\vspace{3pt}
\begin{minipage}{\textwidth}
\footnotesize
\raggedright
\textit{Note.} $-$ denotes abnormal evaluation results.
$^*$ denotes results that are not reported in the original paper.
\end{minipage}
\end{table*}

\section{Experiments}
\label{sec:experiments}
We evaluate ME-VLM on a broad set of benchmarks organized into three groups.
\textbf{Embodied benchmarks} cover four dimensions: physical
understanding, including CV-Bench \cite{tong2024cambrian}, VABench \cite{yuan2025vabench}, RefCOCO \cite{yu2016modeling}, ERQA \cite{team2025gemini}, MindCube \cite{yin2025mindcube}, SPARBench \cite{zhang2026flatland},
SPBench \cite{li2025spatialladderprogressivetrainingspatial}, MMSI-Bench \cite{yang2025mmsi}, ViewSpatial-Bench \cite{li2025viewspatial}, EmbSpatial-Bench \cite{du2024embspatial}, RefSpatial-Bench \cite{zhou2026roborefer}, and
RoboSpatial-Home \cite{song2025robospatial}; task planning, including Cosmos \cite{azzolini2025cosmos}, EgoPlan2 \cite{qiu2024egoplan}, and RoboBench \cite{luo2025robobench};
action execution, including VSIBench \cite{yang2025thinking}, SITE-Bench \cite{wang2025site}, and RoboBench \cite{luo2025robobench}; and error
correction, including RoboFail \cite{sagar2024robofail} and RoboFAC \cite{ye2025robofac}.

\textbf{Agent benchmarks} cover multimodal understanding (BLINK \cite{fu2024blink}, MMStar \cite{chen2024mmstar},
MVBench \cite{li2024mvbench}, VideoMME \cite{fu2024videomme}, RealWorldQA \cite{xai2024realworldqa}), tool and skill invocation (BFCL-V4 \cite{bfcl}, TAU2-Bench \cite{tau-bench}),
long-horizon planning (ClawEval \cite{claweval}), reasoning (GPQA \cite{gpqa}, AIME25 \cite{aime25}, MMLU-Pro \cite{wang2024mmlupro}, MMMU-Pro \cite{yue2024mmmupro},
MathVision \cite{mathvision}), and instruction following (IFEval \cite{ifeval}, IFBench \cite{ifbench}).

\textbf{Task-transfer benchmarks} cover autonomous driving (MAPLM \cite{maplm}, CODA-LM \cite{coda_lm},
DriveAction \cite{driveaction}, nuScenes-QA \cite{Nuscenes-qa}, MME-RealWorld \cite{Mme-realworld}) and embodied navigation (R2R \cite{r2r}, RxR \cite{rxr}).

\begin{table*}[!t]
\caption{\textbf{Results on agent benchmarks. The metrics of the compared methods in this table are re-evaluated under our test framework.}}
\label{tab:multimodal_agent}
\centering
\renewcommand{\arraystretch}{1.25}
\setlength{\tabcolsep}{2pt}
\setlength{\arrayrulewidth}{0.55pt}
\arrayrulecolor{black}
\footnotesize
\begin{tabular}{@{}ll|>{\columncolor{LiLight}}c>{\columncolor{LiLight}}c|cccc@{}}
\hline
\textbf{Subcategory} &
\textbf{Benchmark} &
\makecell{ME-VLM\\4B} &
\makecell{ME-VLM\\35B-A3B} &
\makecell{Hy-Embodied \\30B-A3B} &
\makecell{RynnBrain 1.1 \\ 9B} &
\makecell{PhysBrain 1.5 \\ 8B} &
\makecell{MiMo-Embodied \\ 7B} \\
\hline
\multicolumn{2}{l|}{Average} & 63.1 & 72.5 & 61.2 & 62.9 & 46.6 & 54.1 \\
\hline
\multirow{5}{*}{\makecell[l]{Multimodal\\Understanding}}
& BLINK       & 58.9 & 67.7 & 66.2 & 53.8 & 64.7 & 56.3 \\
& MMStar      & 72.5 & 77.3 & 76.1 & 68.7 & 62.5 & 68.7 \\
& MVBench     & 64.1 & 71.9 & 67.6 & 62.3 & 60.8 & 56.7 \\
& VideoMME    & 55.6 & 60.3 & 58.2 & 55.4 & 52.2 & 25.1 \\
& RealWorldQA & 79.0 & 79.7 & 76.6 & 39.5 & 71.5 & 67.1 \\
\hline
\multirow{2}{*}{\makecell[l]{Long-horizon\\Planning}}
& ClawEval (avg3)                 & -- & 58.0 & -- & 62.4 & -- & 37.0 \\
& ClawEval (pass\textsuperscript{3}) & -- & 31.8 & -- & 37.2 & -- & 6.0 \\
\hline
\multirow{2}{*}{\makecell[l]{Tool or Skill\\Invocation}}
& BFCL-V4     & 40.5 & 63.2 & -- & 63.2 & -- & 51.5 \\
& TAU2-Bench  & 50.1 & 63.9 & -- & 75.6 & -- & 15.0 \\
\hline
\multirow{5}{*}{\makecell[l]{Reasoning}}
& GPQA        & 67.2 & 78.7 & 46.5 & 75.3 & 35.0 & 54.9 \\
& AIME25      & 47.5 & 72.9 & 26.7 & 70.4 & 3.8  & 48.3 \\
& MMLU-Pro    & 75.4 & 82.8 & 74.3 & 81.3 & 46.7 & 71.6 \\
& MMMU-Pro    & 58.4 & 69.3 & 64.8 & 60.8 & 35.1 & 52.1 \\
& MathVision  & 59.6 & 74.3 & 66.1 & 61.4 & 19.6 & 54.5 \\
\hline
\multirow{2}{*}{\makecell[l]{Instruction\\Following}}
& IFEval      & 81.9 & 86.9 & 82.1 & 83.6 & 79.1 & 65.6 \\
& IFBench     & 37.4 & 48.3 & 29.3 & 42.5 & 27.6 & 27.9 \\
\hline
\end{tabular}
\end{table*}

\subsection{Embodied Benchmarks}

\label{sec:experiments:embodied-vlm-benchmarks}

Table~\ref{tab:embodied_benchmarks} presents a comprehensive evaluation of
ME-VLM across 26 embodied benchmarks covering physical understanding, task
planning, action execution, and error correction. ME-VLM~35B-A3B achieves the
highest average score of 70.9, outperforming the strongest competing model,
PhysBrain~1.5-8B, by 8.2 points. It achieves the best score on 14 benchmarks
and ties for the best score on RoboFail-Planning. In physical understanding,
ME-VLM~35B-A3B leads on CV-Bench, VABench-Point, MindCube, SPARBench,
SPBench-MV, ViewSpatial-Bench, EmbSpatial-Bench, and RoboSpatial-Home,
demonstrating strong physical perception, spatial reasoning, and geometric
understanding. Its advantages extend to task planning, action execution, and
failure recovery, where it achieves the best results on Cosmos, EgoPlan2,
SITE-Bench-Video, RoboFail-Execution, and RoboFAC, while matching the best
result on RoboFail-Planning. These results demonstrate the model's ability to
integrate physical perception, spatial reasoning, planning, execution, and
feedback-driven correction within a unified framework.

ME-VLM~4B achieves an average score of 63.4, exceeding all compared baseline
models despite its substantially smaller model capacity. In particular, it
outperforms the strongest baseline average, achieved by PhysBrain~1.5-8B, by
0.7 points. The compact model remains competitive across a broad range of
physical-understanding, task-planning, action-execution, and correction tasks,
including 89.5 on CV-Bench, 67.0 on VABench-Point, 76.6 on RefCOCO-testB, 65.7
on Cosmos, and 79.7 on RoboFail-Execution. The consistently
strong performance of ME-VLM~4B demonstrates that the proposed embodied
post-training framework remains effective at a compact model scale, providing
a favorable balance between embodied capability and deployment efficiency.

\begin{table*}[t]
\caption{\textbf{Results on autonomous driving.}}
\label{tab:Autonomous_Driving}
\centering
\renewcommand{\arraystretch}{1.28}
\setlength{\tabcolsep}{10pt}
\setlength{\arrayrulewidth}{0.55pt}
\arrayrulecolor{black}
\small
\begin{adjustbox}{max width=\textwidth}
\begin{tabular}{@{}l|ccccc@{}}
\hline
\textbf{Method} & MAPLM & CODA-LM &
DriveAction & nuScenes-QA &
MME-RealWorld \\
\hline
Hy-Embodied 30B-A3B & 21.8 & 75.8 & 74.8 & 34.2 & 60.7 \\
RynnBrain 1.1 9B    & 37.5 & 30.5 & 83.2 & 33.7 & 69.0 \\
PhysBrain 1.5 8B    & 34.7 & 56.1 & 63.4 & 29.1 & 47.4 \\
MiMo-Embodied 7B    & 74.5 & 58.6 & 81.0 & 56.7 & 60.3 \\
\hline
\rowcolor{LiLight} ME-VLM 4B       & 41.1 & 70.9 & 75.8 & 54.0 & 66.1 \\
\rowcolor{LiLight} ME-VLM 35B-A3B  & 72.2 & 70.5 & 78.6 & 61.1 & 68.5 \\
\hline
\end{tabular}
\end{adjustbox}

\vspace{3pt}
\begin{minipage}{\textwidth}
\footnotesize
\raggedright
\textit{Note.} Results of MiMo-Embodied are sourced from the original paper;
all other results are evaluated under our test framework.
\vspace{10pt}
\end{minipage}
\end{table*}

\begin{table*}[t]
\caption{\textbf{Results on embodied navigation.}}
\label{tab:nav_comparison}
\centering
\renewcommand{\arraystretch}{1.28}
\setlength{\tabcolsep}{10pt}
\setlength{\arrayrulewidth}{0.55pt}
\arrayrulecolor{black}
\small
\begin{adjustbox}{max width=\textwidth}
\begin{tabular}{@{}l|cccc|cccc@{}}
\hline
\multirow{2}{*}{\textbf{Method}} & \multicolumn{4}{c|}{R2R Val-Unseen} & \multicolumn{4}{c}{RxR Val-Unseen} \\
\cline{2-9}
& NE$\downarrow$ & OS$\uparrow$ & SR$\uparrow$ & SPL$\uparrow$ & NE$\downarrow$ & SR$\uparrow$ & SPL$\uparrow$ & nDTW$\uparrow$ \\
\hline
PhysBrain 1.5 8B & 9.3 & 8.4 & 3.9 & 1.9 & 11.3 & 6.1 & 4.7 & 17.8 \\
RynnBrain 1.1 9B & 9.4 & 15.3 & 5.3 & 1.8 & 11.7 & 4.2 & 2.8 & 15.2 \\
MiMo-Embodied 7B & 9.2 & 9.1 & 2.7 & 0.7 & 11.2 & 3.8 & 3.1 & 15.6 \\
Hy-Embodied 30B-A3B & 9.9 & 14.1 & 5.3 & 4.4 & 11.6 & 8.3 & 5.6 & 26.3 \\
\hline
\rowcolor{LiLight} ME-VLM 4B & 6.4 & 52.1 & 42.9 & 36.8 & 7.5 & 38.0 & 29.3 & 51.2 \\
\rowcolor{LiLight} ME-VLM 35B-A3B & 5.9 & 56.3 & 45.8 & 39.4 & 6.8 & 42.6 & 35.3 & 57.9 \\
\hline
\end{tabular}
\end{adjustbox}
\end{table*}

\subsection{Agent Benchmarks}
\label{sec:experiments:multimodal-agentic-benchmarks}

Table~\ref{tab:multimodal_agent} evaluates ME-VLM across a broad set of
multimodal agent capabilities, including multimodal understanding, tool and
Skill invocation, long-horizon planning, reasoning, and instruction following.
ME-VLM~35B-A3B achieves the highest aggregate score of 72.5, outperforming the
strongest reported baseline, RynnBrain~1.1-9B, by 9.6 points. Its advantage is
particularly consistent in multimodal understanding and reasoning: ME-VLM~35B-A3B
obtains the best results on all five multimodal-understanding benchmarks and all
five reasoning benchmarks. Notable results include 77.3 on MMStar, 71.9 on
MVBench, 60.3 on VideoMME, 78.7 on GPQA, 72.9 on AIME25, 82.8 on MMLU-Pro, and
74.3 on MathVision. It also leads both instruction-following benchmarks,
achieving 86.9 on IFEval and 48.3 on IFBench.

The smaller ME-VLM~4B model remains competitive despite its substantially reduced
model capacity, achieving an aggregate score of 63.1 and slightly outperforming
RynnBrain~1.1-9B, while also exceeding Hy-Embodied~30B-A3B, PhysBrain~1.5-8B,
and MiMo-Embodied~7B. The gap between the two ME-VLM variants is particularly pronounced on tool invocation and complex reasoning. Although
ME-VLM~35B-A3B matches the best result on BFCL-V4, it trails RynnBrain~1.1-9B on
TAU2-Bench and both ClawEval settings, suggesting that highly interactive tool use
and long-horizon agent tasks remain important directions for further improvement.
Overall, the results show that ME-VLM preserves strong general multimodal and
reasoning capabilities while supporting tool use and extended agentic decision
making.

\subsection{Task Transfer}
\label{sec:experiments:task_transfer}
\subsubsection{Autonomous Driving}

Autonomous driving requires models to perform scene understanding, behavioral
reasoning, and decision interpretation in complex and dynamic environments,
making it an important application domain for evaluating the coordination
between embodied cognition and agent capabilities. We adapt ME-VLM to
autonomous-driving tasks and compare it with
representative embodied models on five benchmarks: MAPLM, CODA-LM, DriveAction,
nuScenes-QA, and MME-RealWorld. As shown in
Table~\ref{tab:Autonomous_Driving}, ME-VLM~35B-A3B achieves the highest average
score of 70.2 and demonstrates competitive performance across the evaluated
tasks. It obtains 61.1 on the scene-question-answering benchmark nuScenes-QA,
outperforming all compared baselines, and 68.5 on the real-world
driving-understanding benchmark MME-RealWorld, approaching the best result of
69.0 achieved by RynnBrain 1.1. On MAPLM and DriveAction, ME-VLM~35B-A3B scores
72.2 and 78.6, respectively, remaining competitive with MiMo-Embodied and
RynnBrain 1.1. On CODA-LM, it achieves 70.5, compared with the best result of
75.8 from Hy-Embodied. Overall, these results show that ME-VLM can be adapted to autonomous-driving tasks, demonstrating the generalization potential of its embodied
perception and physical-scene reasoning capabilities.

\begin{figure*}[!t]
    \centering
    \includegraphics[width=0.49\textwidth]{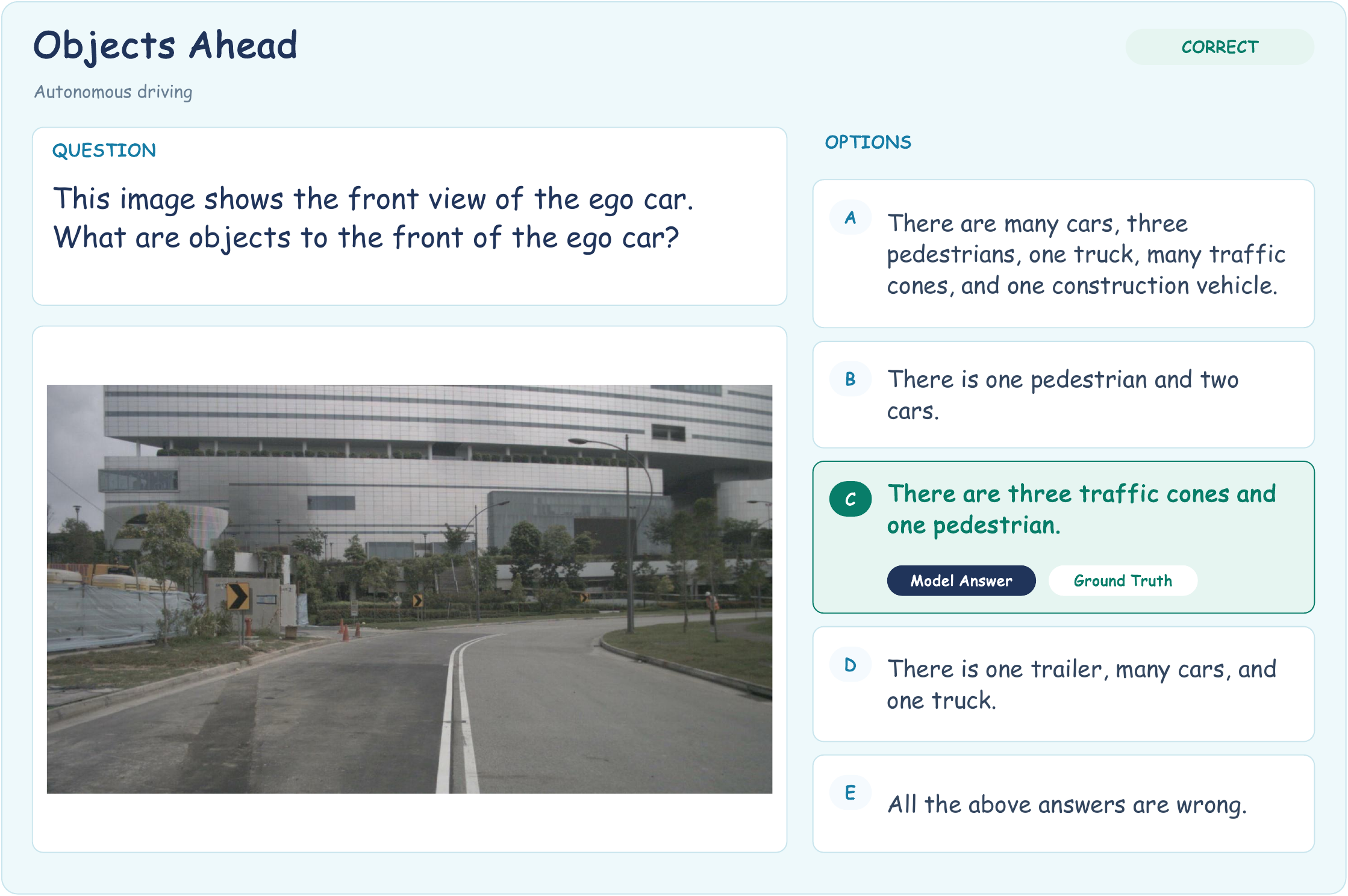}%
    \hfill
    \includegraphics[width=0.49\textwidth]{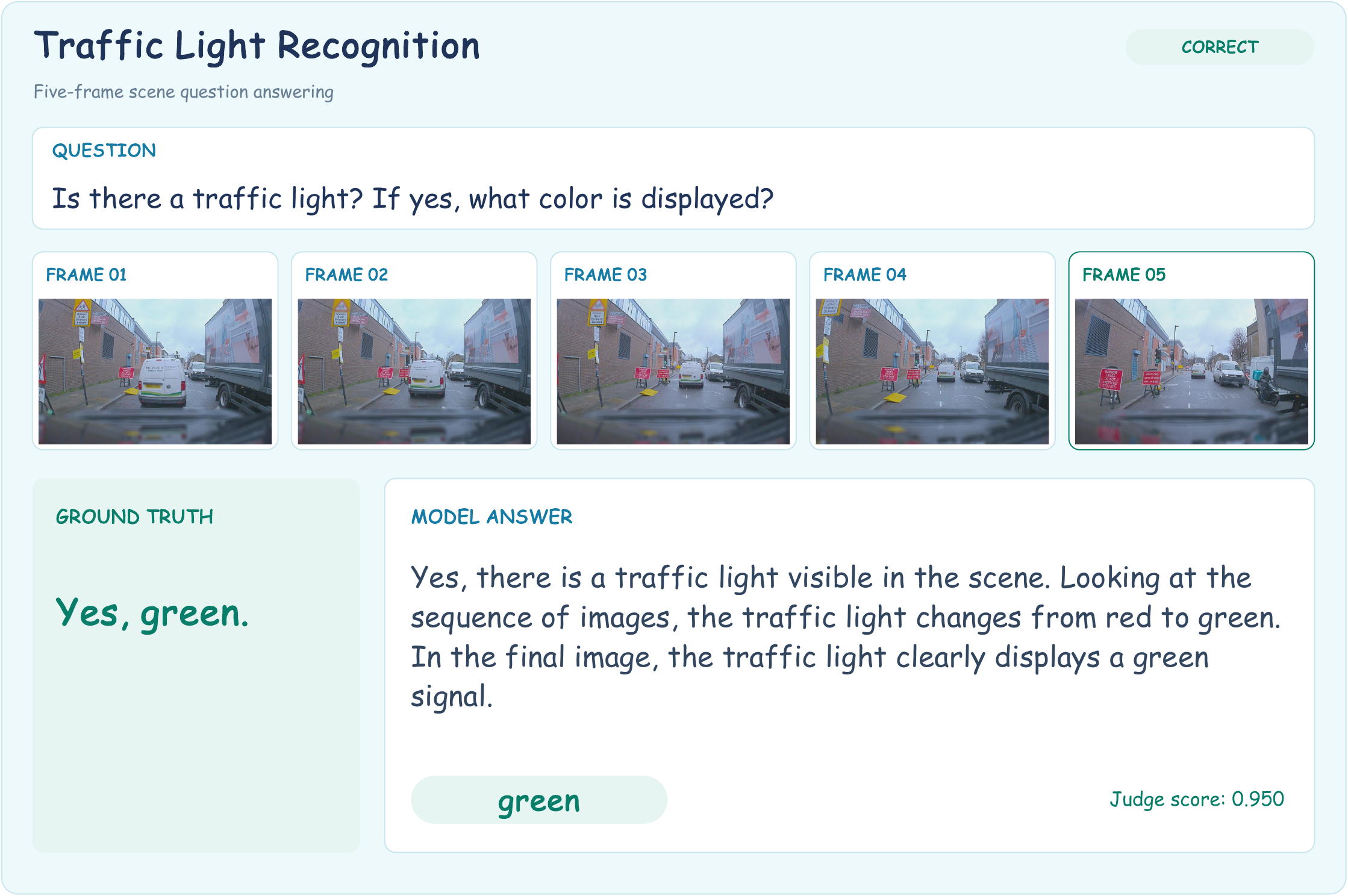}
    \par\medskip
    \includegraphics[width=0.49\textwidth]{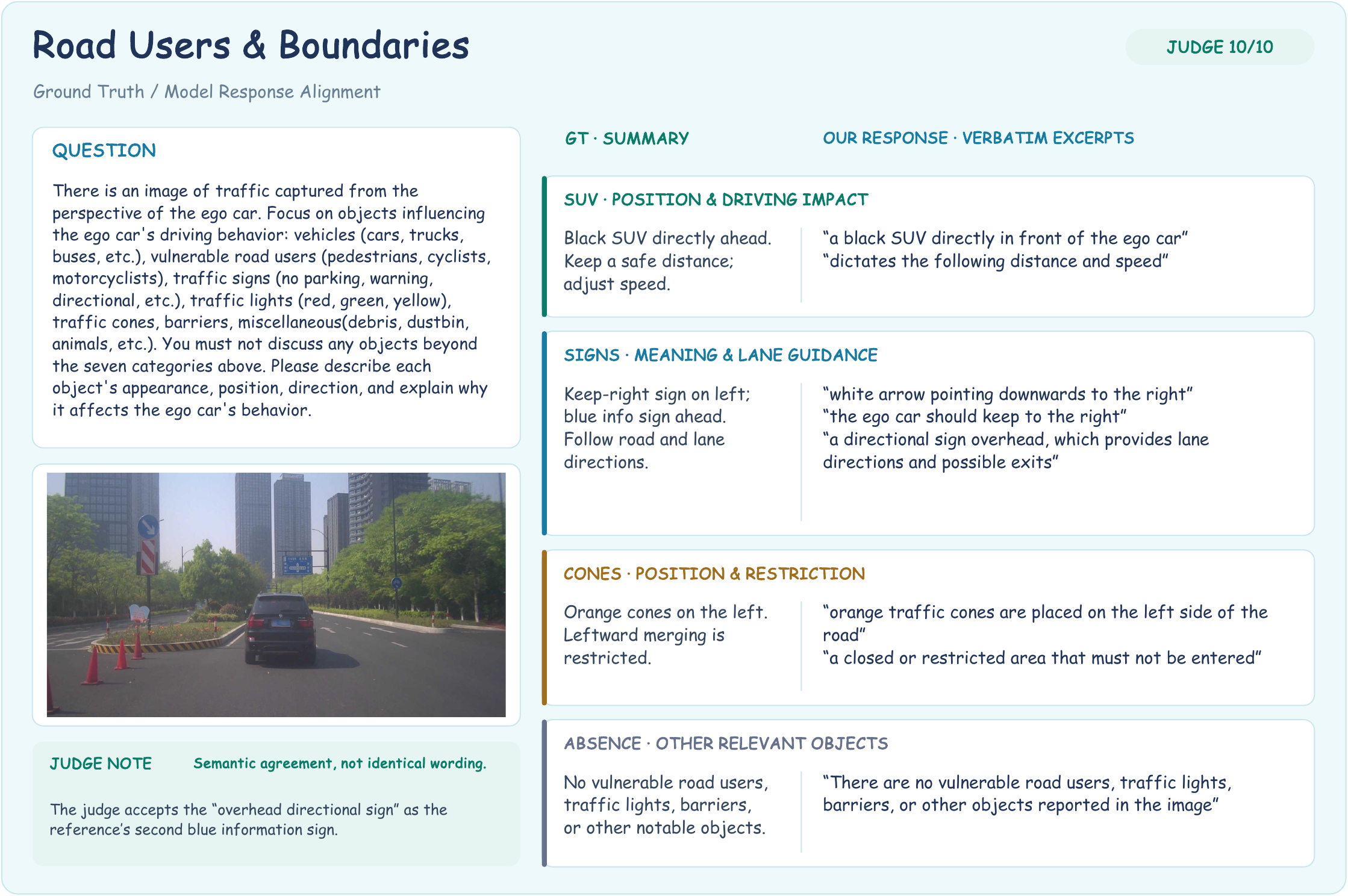}%
    \hfill
    \includegraphics[width=0.49\textwidth]{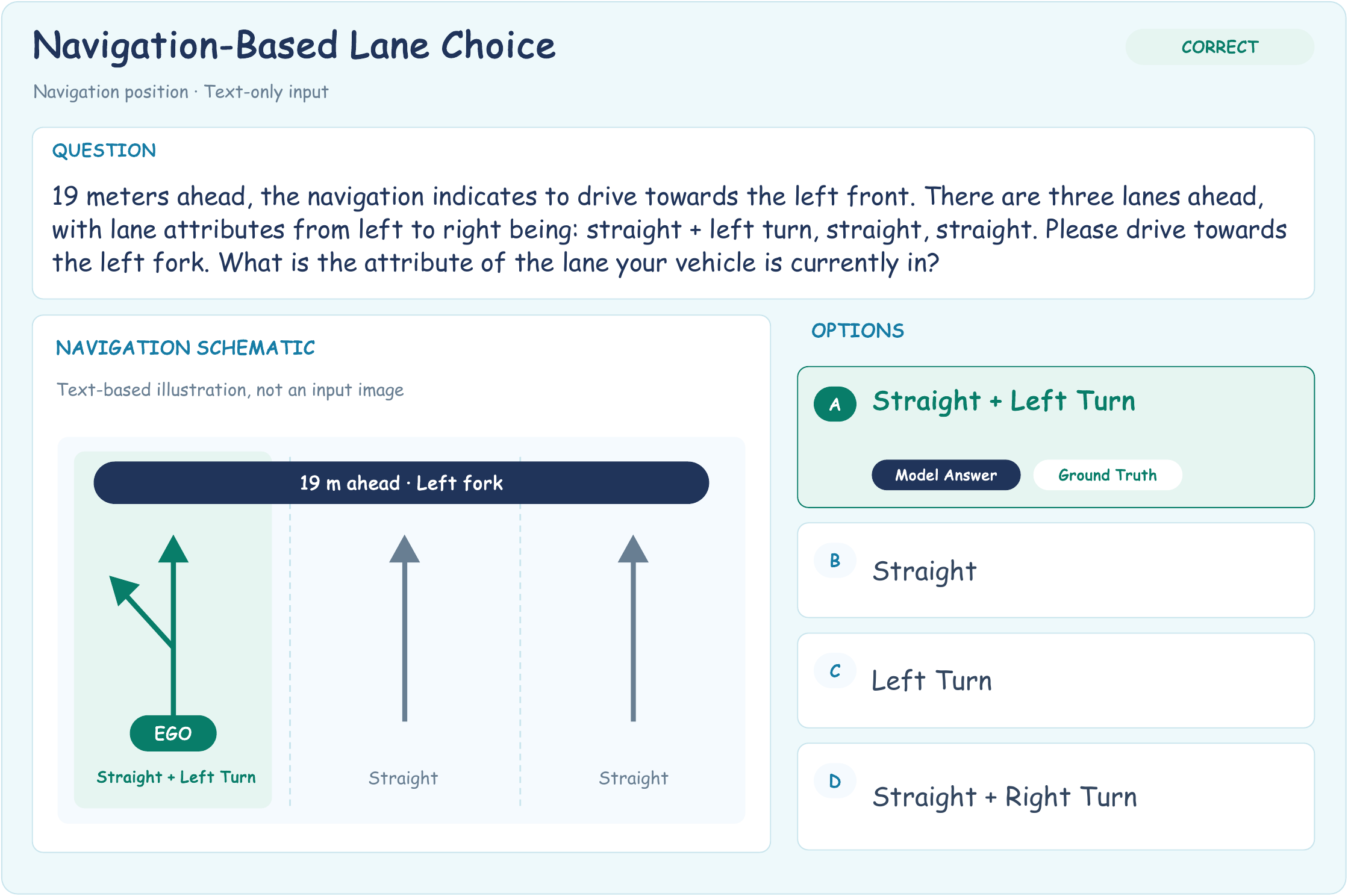}
    \caption{\textbf{Autonomous-driving cases.} Qualitative examples of traffic-scene understanding, temporal state recognition, decision reasoning, and lane-intent inference.}
    \label{fig:ad-case}
\end{figure*}

\subsubsection{Embodied Navigation}

Embodied navigation requires an agent to follow natural-language instructions
and perform path planning and long-horizon execution in three-dimensional
environments, placing substantial demands on spatial perception, instruction
grounding, and cross-step state tracking. We evaluate ME-VLM on the R2R
Val-Unseen and RxR Val-Unseen benchmarks against representative embodied
models, including PhysBrain~1.5, RynnBrain~1.1, MiMo-Embodied, and
Hy-Embodied. As shown in Table~\ref{tab:nav_comparison}, ME-VLM~35B-A3B
substantially outperforms all competing models across both benchmarks. On R2R
Val-Unseen, it achieves a navigation error (NE) of 5.9, an oracle success rate
(OS) of 56.3, a success rate (SR) of 45.8, and a success rate weighted by path
length (SPL) of 39.4. In comparison, the strongest baseline results are 9.2 in
NE, 15.3 in OS, 5.3 in SR, and 4.4 in SPL. On RxR Val-Unseen, ME-VLM~35B-A3B
achieves an NE of 6.8, an SR of 42.6, an SPL of 35.3, and an nDTW score of
57.9. These results considerably exceed the strongest baseline scores of 11.2
in NE, 8.3 in SR, 5.6 in SPL, and 26.3 in nDTW. The compact ME-VLM~4B model
also consistently surpasses the compared baselines, demonstrating that the
improvement is preserved across model scales. These results indicate that the
spatial reasoning, temporal state modeling, and long-horizon planning
capabilities developed during embodied post-training can be effectively adapted
to navigation, enabling strong coordination between language-instruction
understanding and path execution.


\section{Applications}
\label{sec:applications}

\begin{figure*}[!t]
    \centering
    \includegraphics[width=0.49\textwidth]{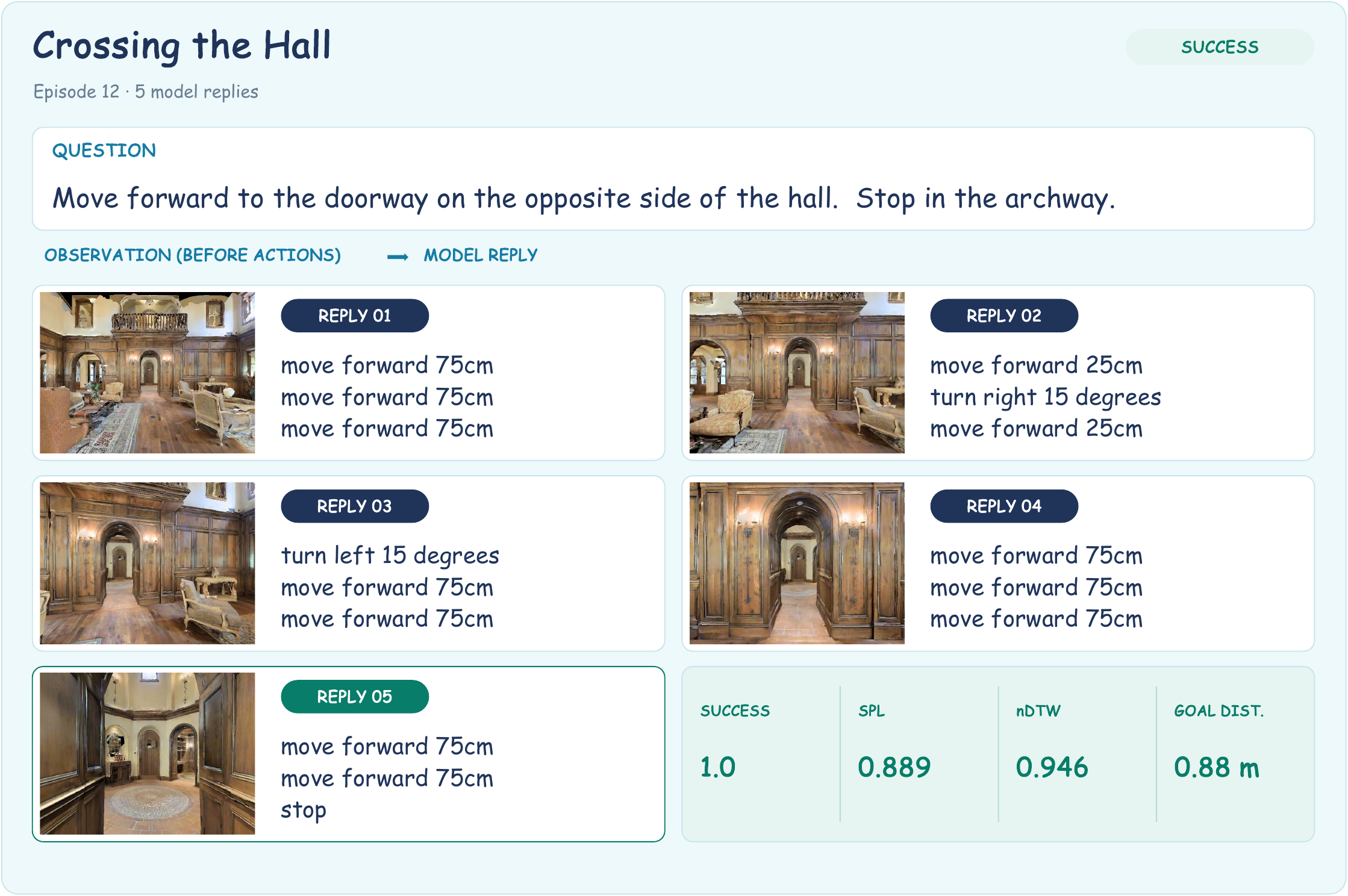}%
    \hfill
    \includegraphics[width=0.49\textwidth]{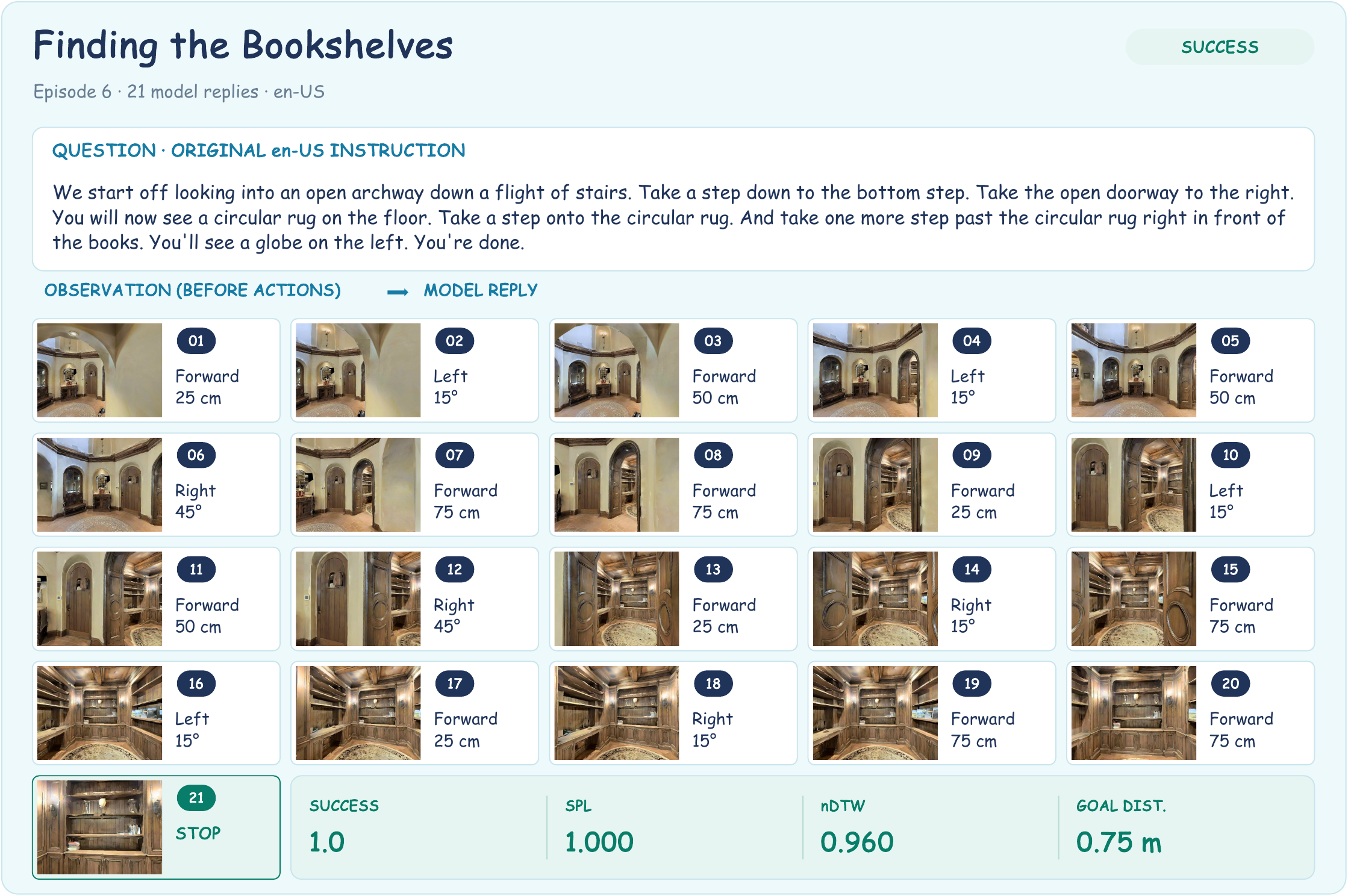}
    \caption{\textbf{Embodied-navigation cases.} Instruction-following navigation with multi-action and single-action response formats.}
    \label{fig:nav-case}
\end{figure*}

We further examine how the capabilities of ME-VLM generalize to representative applications spanning embodied task execution, digital tool use, and closed-loop interaction with physical environments. The following qualitative cases emphasize how the model grounds instructions in observations, organizes multi-step actions, incorporates execution feedback, and produces verifiable task outcomes.

\subsection{Embodied Applications}
\label{sec:applications:embodied-domain}

\subsubsection{Autonomous Driving}
\label{sec:applications:autonomous-driving}

Autonomous driving requires a model to jointly understand dynamic traffic scenes, track temporally evolving states, and relate perceived objects to driving decisions. Figure~\ref{fig:ad-case} presents four cases covering complementary aspects of these capabilities. ME-VLM identifies traffic cones and pedestrians ahead, tracks a green traffic signal across successive observations, and explains how the preceding vehicle, road signs, and traffic cones constrain the appropriate driving behavior. The text-only case further shows that the model can combine a navigation instruction with lane descriptions to infer that the current lane permits both through and left-turn movements. Across these cases, the generated responses remain semantically consistent with the reference answers, demonstrating reliable scene understanding and driving-oriented reasoning under diverse input formats.


\subsubsection{Embodied Navigation}
\label{sec:applications:embodied-navigation}

Embodied navigation requires the model to ground natural-language instructions in a sequence of visual observations while maintaining spatial and execution states over long horizons. Figure~\ref{fig:nav-case} shows two successful navigation cases, where each image corresponds to the observation preceding the associated response. In the hall-traversal task, ME-VLM reaches the goal in five responses, each containing three actions that combine forward motion with orientation adjustment. In the longer landmark-based task, the model passes through a doorway and reaches the target bookshelf in 21 single-action responses. The final distances to the two goals are 0.88~m and 0.75~m, respectively. These cases show that ME-VLM can support both compact multi-action prediction and step-by-step control while preserving instruction grounding and spatial consistency throughout execution.

\begin{figure*}[!t]
    \centering
    \includegraphics[width=\textwidth]{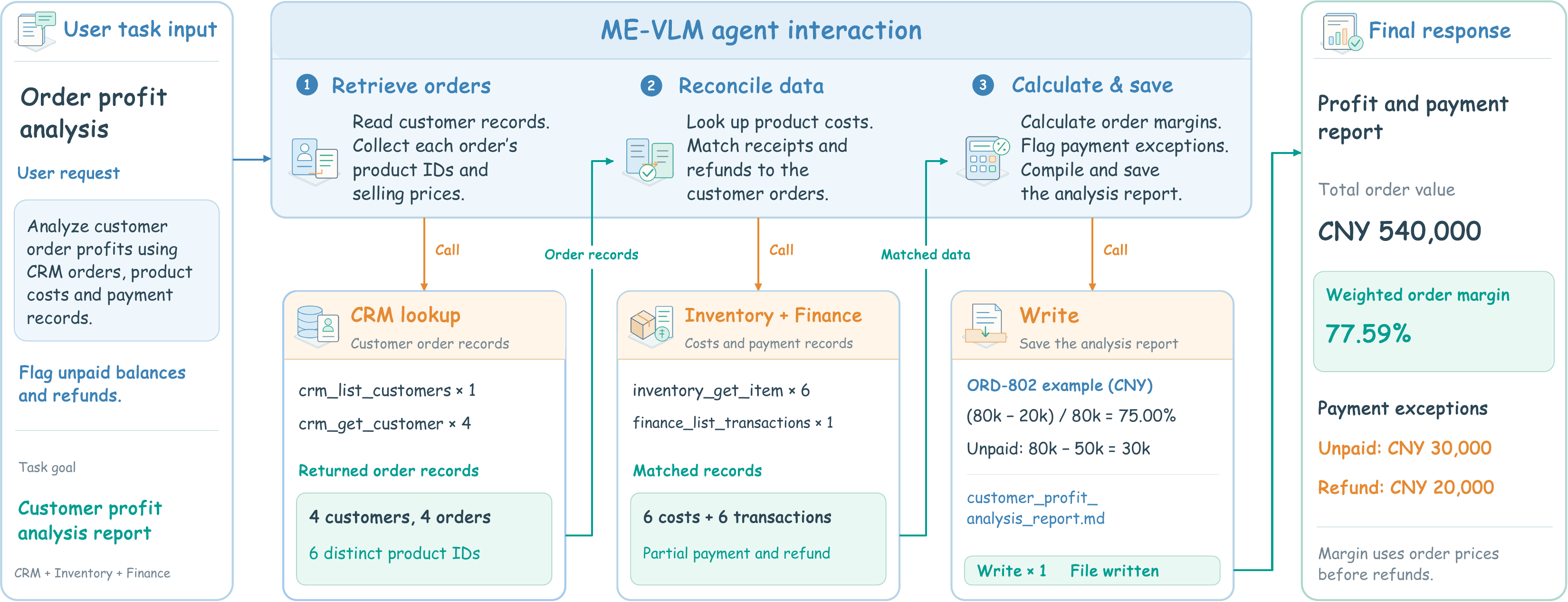}
    \caption{\textbf{Order-profit analysis.} ME-VLM retrieves and reconciles enterprise records, calculates profit statistics, identifies payment exceptions, and writes the final report.}
    \label{fig:agent_case_order_profit}
\end{figure*}
\begin{figure*}[!t]
    \centering
    \includegraphics[width=\textwidth]{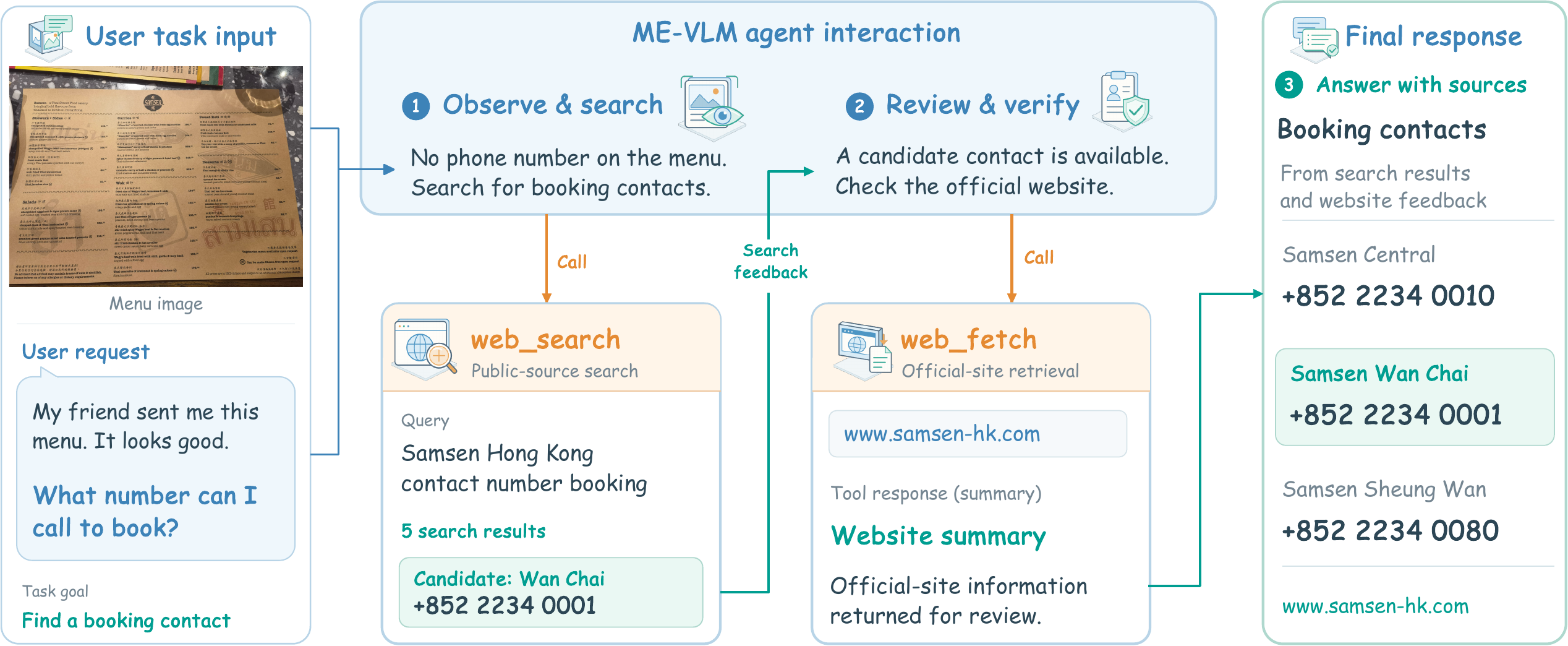}
    \caption{\textbf{Visually grounded contact retrieval.} ME-VLM searches for information absent from the input image, verifies candidate contacts against the official website, and answers with supporting evidence.}
    \label{fig:agent_case_booking_contact}
\end{figure*}
\subsection{Digital Agent Applications}
\label{sec:applications:digital-agent}

The following cases examine whether ME-VLM can extend its perception and planning capabilities to multi-step tasks involving external information and digital tools. In both cases, the model must identify the information required by the user, invoke appropriate tools, incorporate intermediate results, and return a verifiable final response.

\paragraph{Order Profit Analysis.}
Figure~\ref{fig:agent_case_order_profit} presents a business-analysis task involving multiple enterprise systems and structured records. Given a high-level request, ME-VLM retrieves customer orders from the CRM system, extracts product identifiers and selling prices, and queries inventory and finance tools for the corresponding costs and transaction records. It then reconciles the retrieved information, calculates order-level margins, identifies unpaid balances and refunds, and writes the complete analysis to a report. The case shows that ME-VLM can decompose an open-ended objective, coordinate successive tool calls, maintain intermediate task states, and consolidate execution results into a structured and verifiable output.

\paragraph{Visually Grounded Contact Retrieval.}
Figure~\ref{fig:agent_case_booking_contact} shows a case that combines image understanding with external information retrieval. Given a menu image and a request for a booking phone number, ME-VLM identifies the restaurant while recognizing that the requested contact information is absent from the image. It therefore searches for candidate contacts and retrieves the official website to verify the results before answering. The final response organizes the verified phone numbers by branch and provides the supporting source. This case demonstrates the model's ability to recognize insufficient visual evidence, select appropriate tools, verify retrieved information, and produce an evidence-grounded response rather than relying on an unsupported inference.

\subsection{Embodied Agent Applications}
\label{sec:applications:embodied-agent}

Beyond perception and open-loop action prediction, an embodied agent must connect natural-language objectives to executable skills and revise its decisions according to changes in the physical state. We present simulation and real-world manipulation cases to examine spatial grounding, action sequencing, execution monitoring, and feedback-driven correction within this closed interaction loop.

\subsubsection{Simulation Environment}
\label{sec:applications:simulation-environment}

The simulation cases examine how ME-VLM translates visual and linguistic observations into manipulation actions under controlled execution feedback. The first case focuses on spatial reference resolution and single-round planning, whereas the second evaluates recovery from initial perception and execution errors.

\begin{figure}[!htbp]
    \centering
    \includegraphics[width=0.49\textwidth]{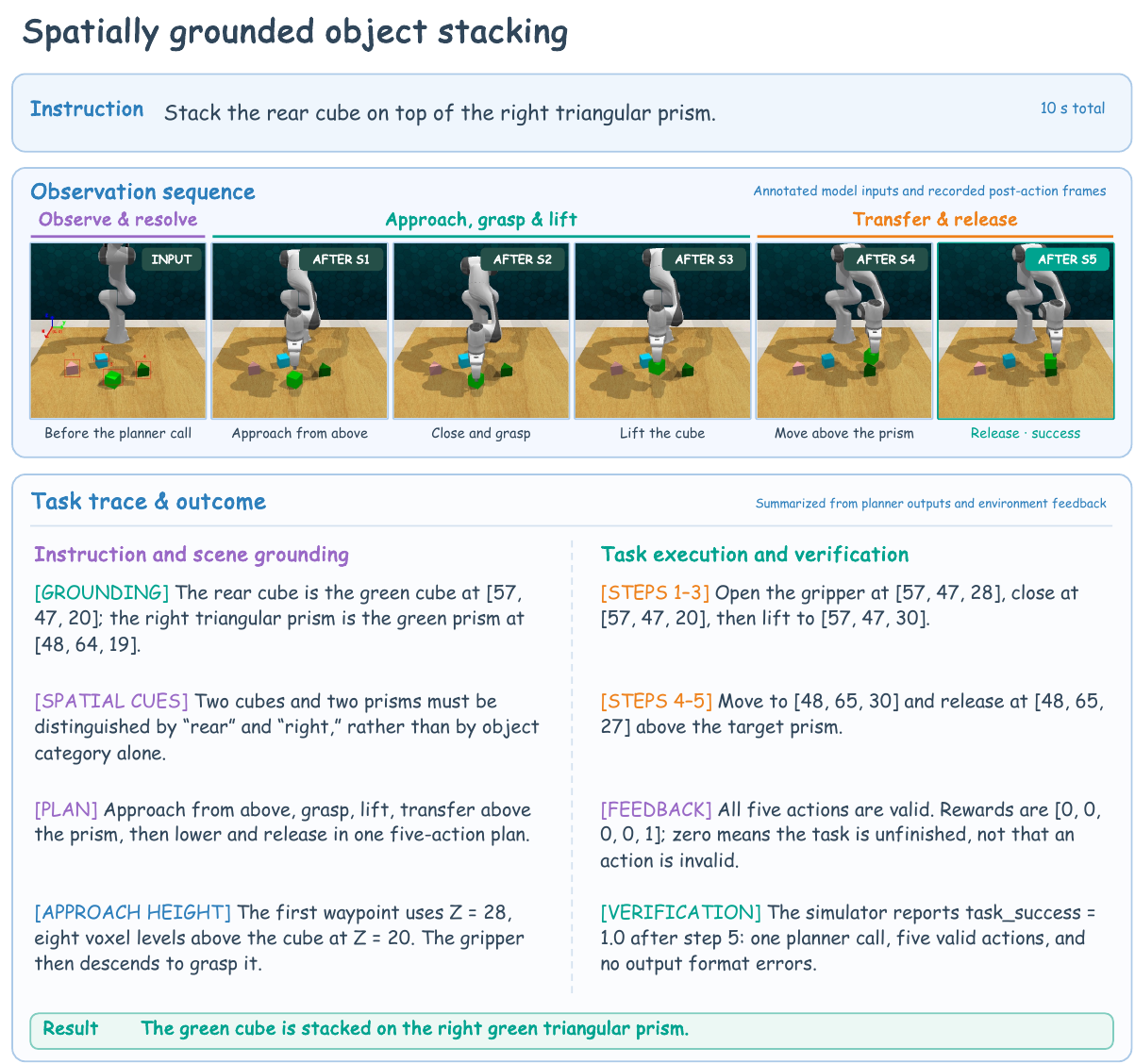}%
    \hfill
    \includegraphics[width=0.49\textwidth]{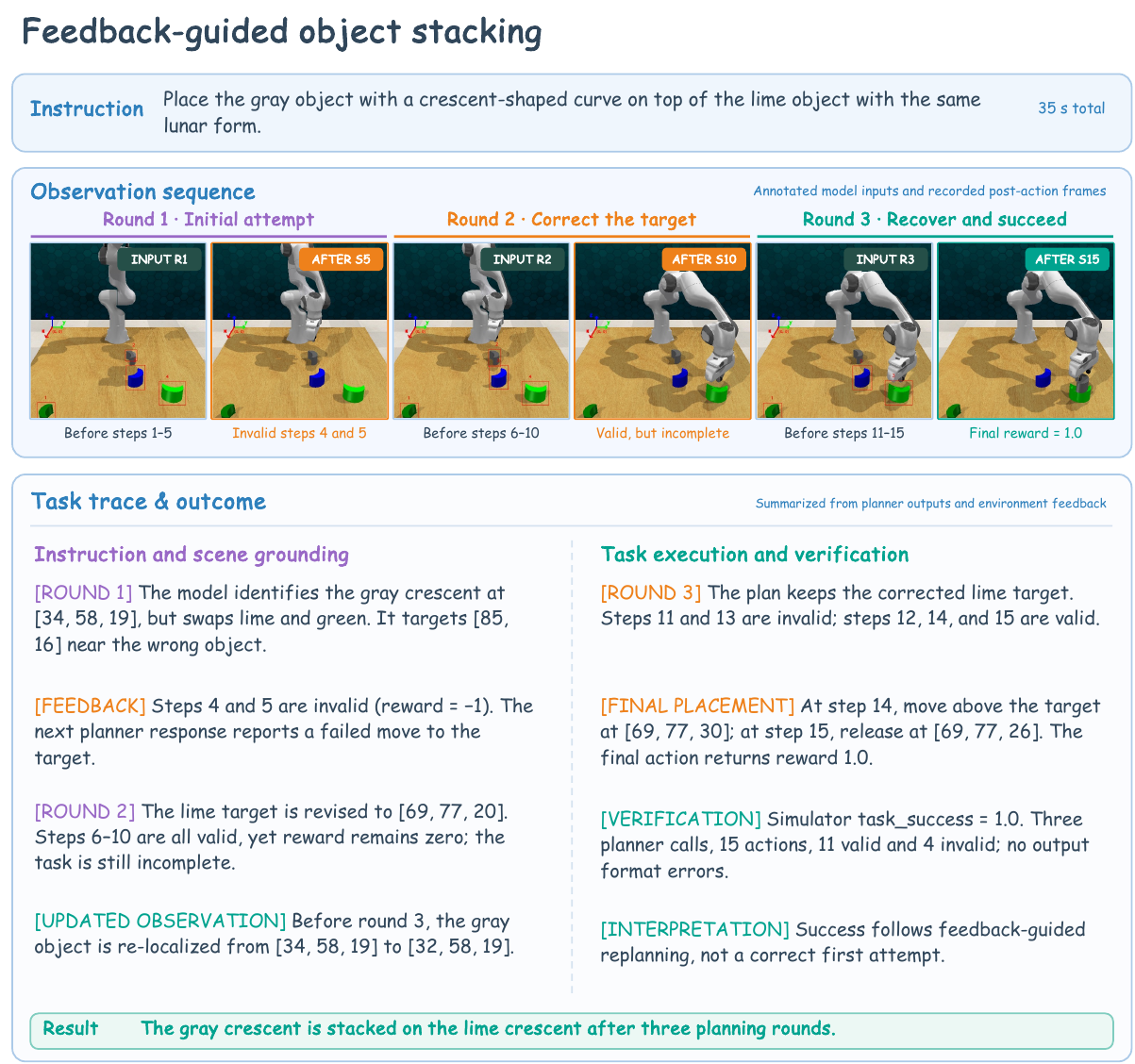}
    \caption{\textbf{Simulated object stacking.} Left: ME-VLM resolves spatial references and completes the task with five valid actions in a single planning round. Right: ME-VLM revises its target identification and action plan using execution feedback, completing the task after three planning rounds.}
    \label{fig:applications:sim-world-case}
\end{figure}

\paragraph{Single-Round Object Stacking.}
In the left panel of Figure~\ref{fig:applications:sim-world-case}, the model receives the instruction ``Stack the cube at the back on the triangular prism on the right.'' It resolves the spatial references to identify the green cube and the rightmost green triangular prism among multiple objects of the same types. ME-VLM then produces an ordered sequence of approach, grasp, lift, transfer, and release actions. The simulator confirms successful stacking after five valid actions in a single planning round, showing that the model can connect spatially grounded object recognition to an executable manipulation plan.

\paragraph{Feedback-Driven Error Correction.}
In the right panel of Figure~\ref{fig:applications:sim-world-case}, the instruction is ``Place the gray crescent-shaped object on the lime-colored object of the same shape.'' The model initially confuses the lime-colored target with a green object, leading to invalid actions during execution. After receiving environment feedback and an updated visual observation, it revises the target identity and placement coordinates while accounting for the changed position of the manipulated object. The task is completed after three planning rounds. Rather than relying on a correct initial prediction, this case highlights the model's ability to incorporate execution feedback, update its scene state, and iteratively recover from perception and action errors.

\subsubsection{Real-World Environment}
\label{sec:applications:realworld-environment}

Real-world embodied tasks additionally require the model to align user intent with changing physical observations and to determine both what actions to execute and when to execute them. The following robotic manipulation cases evaluate compound-instruction decomposition and visually triggered execution in a physical environment.

\begin{figure}[!htbp]
    \centering
    \includegraphics[width=0.49\textwidth]{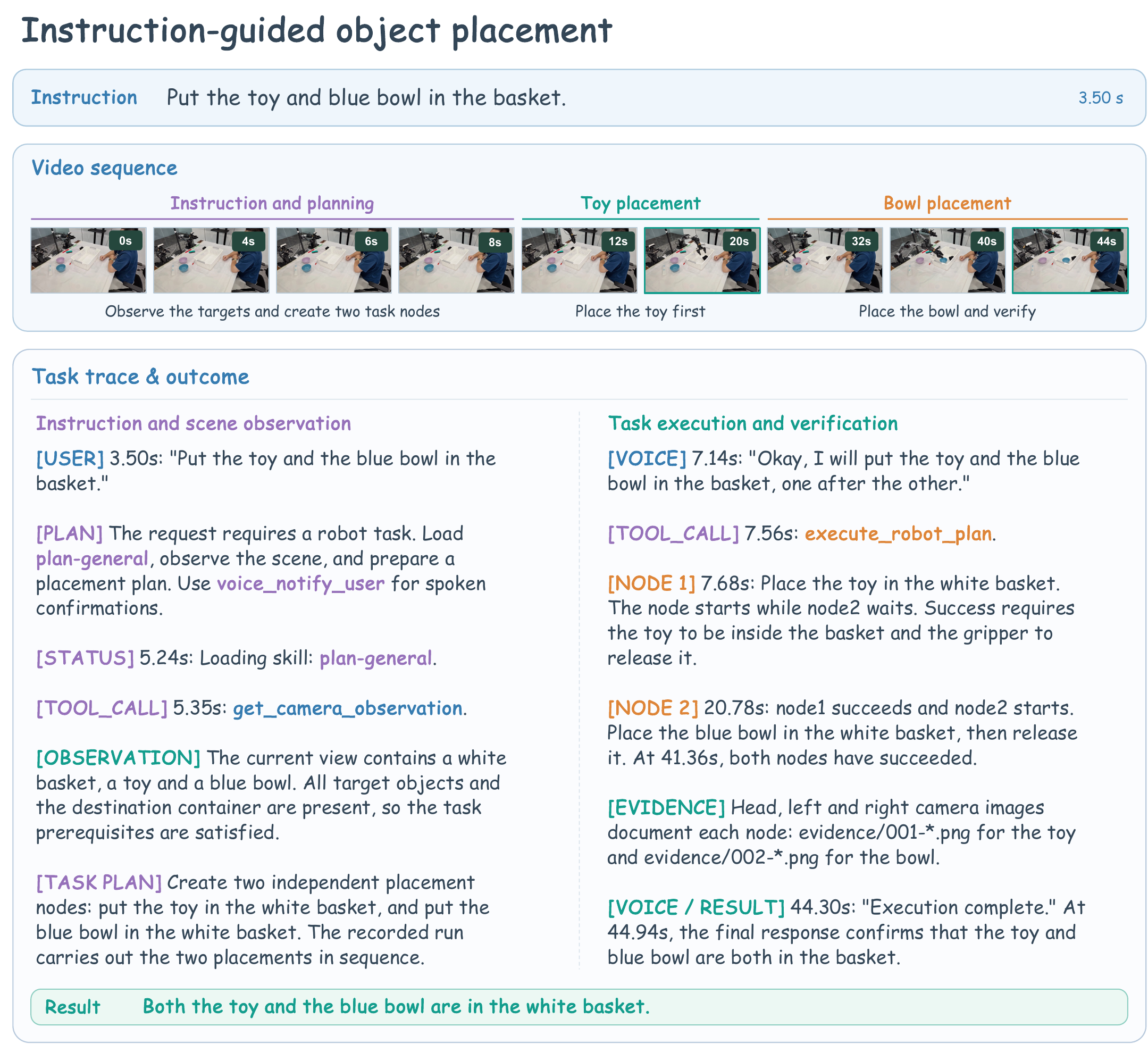}%
    \hfill
    \includegraphics[width=0.49\textwidth]{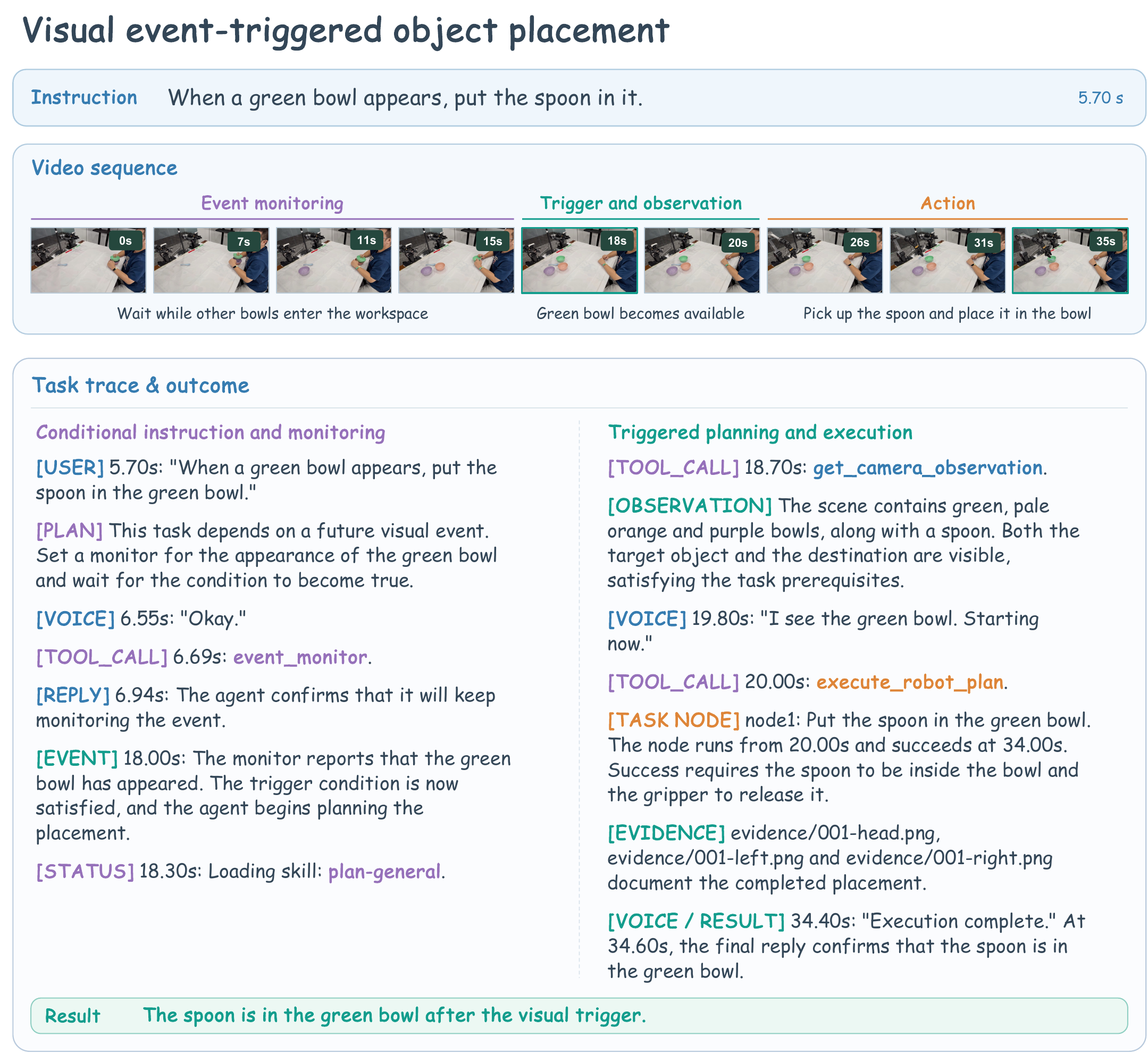}
    \caption{\textbf{Real-world robotic manipulation.} Left: ME-VLM decomposes a multi-object instruction and sequentially places both targets into the basket. Right: ME-VLM monitors the environment and executes the requested placement after verifying that the specified visual condition is satisfied.}
    \label{fig:applications:realworld-case}
    \vspace{-5pt}
\end{figure}

\paragraph{Compound Object Placement.}
In the left panel of Figure~\ref{fig:applications:realworld-case}, the user instructs the system to ``Put the toy and blue bowl in the basket.'' ME-VLM identifies the two target objects and the shared destination, then decomposes the instruction into two placement subtasks. The system places the toy first and the blue bowl second, successfully completing both operations through sequential Skill invocation. This case shows how ME-VLM converts a compound instruction into a structured execution plan while maintaining the target and destination states across multiple physical actions.

\paragraph{Visual Event-Triggered Placement.}
In the right panel of Figure~\ref{fig:applications:realworld-case}, the user provides the conditional instruction ``When a green bowl appears, put the spoon in it.'' ME-VLM interprets the appearance of the green bowl as the execution condition and monitors the scene until the event occurs. It then uses a new visual observation to verify the target and scene state before invoking the placement skill. This case illustrates the coordination of conditional-instruction understanding, visual event monitoring, scene verification, and physical execution, enabling the model to initiate an action at the appropriate time rather than immediately after receiving the instruction.

\section{Conclusion}
\label{sec:conclusion}

In this work, we presented ME-VLM, a unified embodied VLM for physical perception, spatial reasoning, task planning, and feedback-driven execution. ME-VLM achieves strong performance on embodied benchmarks, and its capabilities acquired through the unified training pipeline can be directly applied to autonomous driving and embodied navigation. Experiments in simulated and real-world environments further demonstrate its ability to ground instructions in physical scenes, execute multi-step actions, respond to environmental changes, and recover from failures using execution feedback. Moreover, visual token compression, W4A8 quantization, and hardware-software co-design enable local deployment of ME-VLM 4B on the M100 NPU with controlled accuracy degradation, providing a practical foundation for closed-loop embodied intelligence in real-world environments.

\clearpage
\bibliography{references}

\clearpage

\section{Contributions}
\label{sec:contributions}
\subsection*{Core Contributors}

\subsubsection*{Model:}
Xuhan Zhu, Maokui He, Zide Liu, Chunpeng Zhou,  Xianwei Mao, Wei He, Chenfeng Wang, Hengtao Li, Shengyu Yao, Chang Ren, Chaoqun Du, Zeyu Zhang, Shuai Guo, Fan Lu, Xiyue Zhang, Lifu Mu, Chenrui Cui, Jia Shi, Haipeng Liu, Huazhao Kang, Yuying Chen

\subsubsection*{Deployment:}
Li Zeng, Cheng Qian, Zhiyuan Man, Shiqin Lin, Peng Zhou, Zhenyang Wang

\subsubsection*{Project Leaders:}
Pengfei Yu, Ning Mao, Zhichao Wang

\subsubsection*{Advisors:}
Yan Xie, Kun Zhan, Pan Zhou, Yu Liu

\subsection*{Contributors}
Bing Zhang, Rong Li, Bolong Tian, Xiaoke Ming, Baolan Gao, Qiang Zhang, Xin Wang, Hongsheng Xin, Huimin Ren, Chuanyu Han, Xueyang Zhang, Danlu Dong, Jiankun Qi, Mofan Zhou, Chen Lu, Qifang Wu


\clearpage

\end{document}